\documentclass[journal]{IEEEtran}
\usepackage{amsmath,amsfonts}
\usepackage{algorithmic}
\usepackage{algorithm}
\usepackage{array}
\usepackage{balance}
\usepackage{booktabs}
\usepackage[table]{xcolor}  
\usepackage{bm}
\usepackage[caption=false,font=normalsize,labelfont=sf,textfont=sf]{subfig}
\usepackage{makecell}
\usepackage{textcomp}
\usepackage{stfloats}
\usepackage{url}
\usepackage{tikz}

\usepackage{verbatim}
\usepackage{graphicx}

\usepackage{ragged2e}
\usepackage{stfloats}
\usepackage[numbers]{natbib}
\usepackage[colorlinks=true,
linkcolor=blue,citecolor=blue,urlcolor=blue]{hyperref}

\newcommand\copyrighttext{%
  \footnotesize \textcopyright 2025 IEEE. Personal use of this material is permitted.
  Permission from IEEE must be obtained for all other uses, in any current or future
  media, including reprinting/republishing this material for advertising or promotional
  purposes, creating new collective works, for resale or redistribution to servers or
  lists, or reuse of any copyrighted component of this work in other works.
  DOI: \href{https://ieeexplore.ieee.org/document/11089993}{10.1109/TNNLS.2025.3585949}}
\newcommand\copyrightnotice{%
\begin{tikzpicture}[remember picture,overlay]
\node[anchor=south,yshift=4pt] at (current page.south) {\fbox{\parbox{\dimexpr\textwidth-\fboxsep-\fboxrule\relax}{\copyrighttext}}};
\end{tikzpicture}%
}

\hypersetup{
  pdfinfo={
  pdfproducer={},
  Title={},
  Subject={},
  Author={},
  }
}

\begin{document}

\title{Video Prediction of Dynamic Physical Simulations with Pixel-Space Spatiotemporal Transformers}

\author{Dean L. Slack, G. Thomas Hudson, Thomas Winterbottom, Noura Al Moubayed\\
Durham University, UK\\
\texttt{{dean.l.slack@durham.ac.uk}}
}




\maketitle
\copyrightnotice

\vspace{-1mm}
\begin{abstract}
Inspired by the performance and scalability of autoregressive large language models, transformer-based models have seen recent success in the visual domain. This study investigates a transformer adaptation for video prediction with a simple end-to-end approach, comparing various spatiotemporal self-attention layouts. Focusing on causal modelling of physical simulations over time; a common shortcoming of existing video-generative approaches, we attempt to isolate spatiotemporal reasoning via physical object tracking metrics and unsupervised training on physical simulation datasets. We introduce a simple yet effective pure transformer model for autoregressive video prediction, utilising continuous pixel-space representations for video prediction. Without the need for complex training strategies or latent feature-learning components, our approach significantly extends the time horizon for physically accurate predictions by up to 50\% when compared with existing latent-space approaches, while maintaining comparable performance on common video quality metrics.  Additionally, we conduct interpretability experiments to identify network regions that encode information useful to perform accurate estimations of PDE simulation parameters via probing models, and find this generalises to the estimation of out-of-distribution simulation parameters. This work serves as a platform for further attention-based spatiotemporal modelling of videos via a simple, parameter-efficient, and interpretable approach.
\end{abstract}

\begin{IEEEkeywords}
Video prediction, spatiotemporal transformers, pixel-space modelling, physics modelling, autoregressive models, hierarchical video transformers.
\end{IEEEkeywords}

\section{Introduction}
\IEEEPARstart{R}{ecent} progress in the development of transformer \citep{vaswani2017attention} based generative models, particularly text-generative models in Natural Language Processing (NLP), have led to increased efforts to extend their application beyond the linguistic domain \citep{dosovitskiy2021an, yan2021videogpt, oprea2020review}. Building on the success of generative modelling in the image domain, such as Variational Autoencoders (VAEs) \citep{razavi2019generating} and Diffusion models \citep{zhang2023text}, recent advances have extended to generative modelling of videos. This is becoming an area of increasing research, focusing on the development of novel architectures and techniques for model interpretability \citep{castrejon2019improved,oprea2020review,zhou2020deep}. In this work we investigate both - taking direct inspiration from the performance and scalability of Large Language Models (LLMs), we investigate a pure transformer model as an end-to-end approach for unsupervised video prediction, focusing on physical simulation datasets driven by partial differential equations (PDEs) as a quantifiable measure of spatiotemporal reasoning. Our Pixel-Space Spatiotemporal Video Transformer (PSViT) offers a highly simplified approach for end-to-end video prediction while extending the time horizon of physically accurate outputs.
\IEEEpubidadjcol
Autoregressive transformer LLMs at scale have been shown to exhibit emergent properties beyond their apparent pretraining goals \citep{wei2022emergent, Brown2020LanguageMA}. Video generation is therefore a natural next step for causal modelling for both input complexity and computational demand, requiring innovation and adaptations of existing techniques in deep generative modelling. The emergent and highly generalisable behaviour of autoregressive LLMs hints at promising applications of spatiotemporal modelling beyond conditional video generation, particularly where underlying laws are not directly observable from image pixel-space, such as simulating fluid dynamics \citep{kohl2023_acdm}, weather forecasting \citep{sønderby2020metnet}, robot motion planning \citep{7989324}, future scenarios for autonomous driving \citep{wen2023panacea, hu2023gaia1generativeworldmodel}, and traffic prediction \citep{Gao_2022_CVPR}.

\begin{figure}[t!]
    \centering 
    \includegraphics[width=0.48\textwidth]{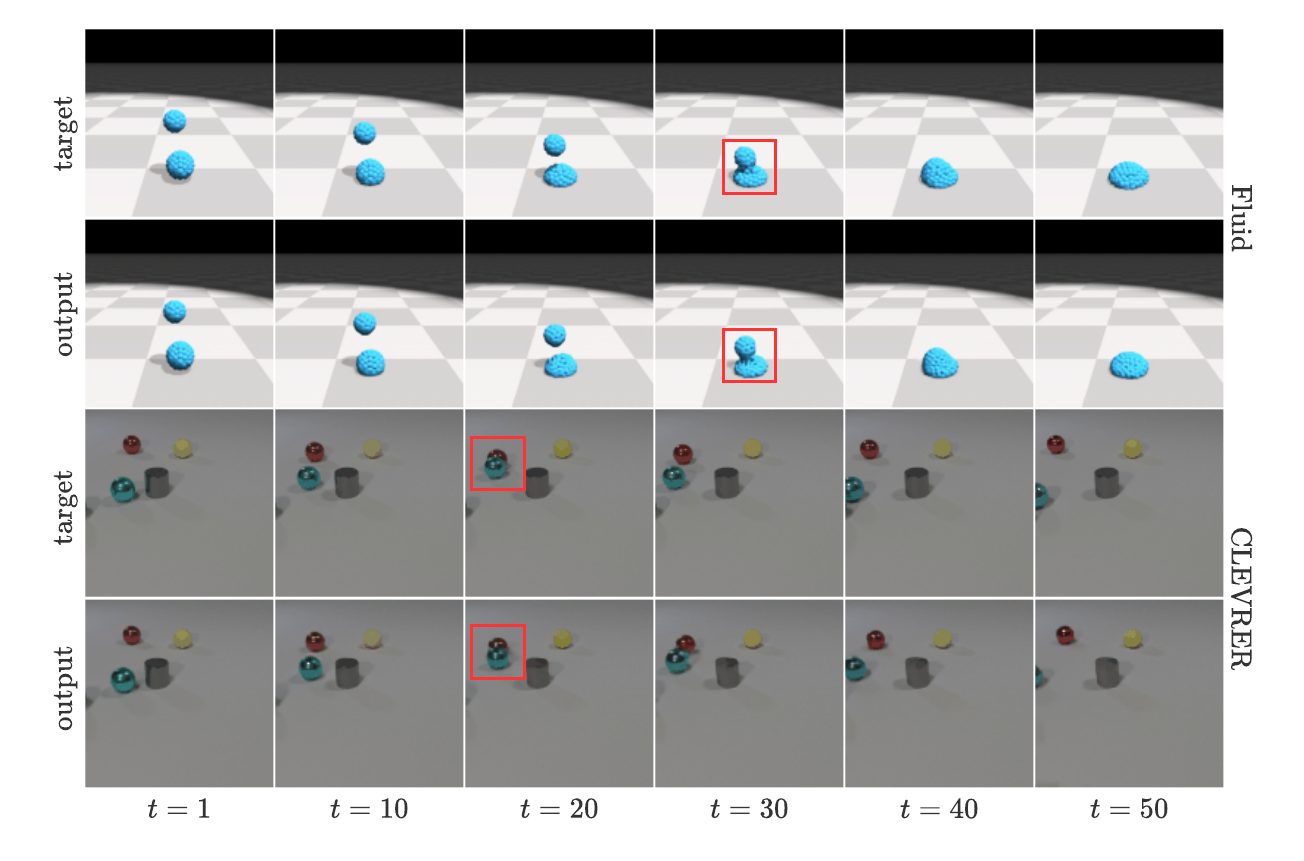}\\
    \caption{Example outputs from our PSViT model trained on video simulation data exhibiting physical dynamics. Successful collision prediction time-step annotated by the red boxes.}
    \label{fig1}
\vspace{-15pt}
\end{figure}

Most existing video generative models are evaluated using pixel-based or perceptual quality metrics, conditioned on prior frames or textual prompts, and often produce stochastic outputs \citep{xing2023survey, croitoru2023diffusion, yu2023magvit}. Such approaches typically employ an \emph{encoder-predictor-decoder} architecture to learn a compressed latent representation of frames, with a predictor backbone enforcing causality for future frame generation. However, there is little research studying physical accuracy of generated videos over time, a known failing of these models \citep{ming2024survey}. This gap motivates our investigation into whether modelling continuous pixel-space representations offers a simpler path to improved, physically coherent video prediction. Consequently, we explore a simple, effective, and interpretable autoregressive transformer for end-to-end video prediction (see Fig.~\ref{fig1} for examples). Our primary focus is the modelling of videos depicting physical simulations, enabling quantitative evaluation of learned physical dynamics via object positioning over time and comparison with state-of-the-art methods. We propose that continuous pixel-space modelling is a viable strategy for developing transformers as simple, interpretable end-to-end video prediction models. Furthermore, our experiments investigate the extent to which our model encodes sequence-specific PDE parameters governing the physical simulations. Our key contributions can be summarised as follows:
\begin{itemize}
    \item We propose a novel, simple, and effective end-to-end transformer with a U-Net style architecture for autoregressive video prediction. This model omits complex architectural priors or training goals, and investigates various patch-wise spatiotemporal self-attention strategies.
    \item Using an object tracking metric, we demonstrate that our model achieves increased temporal accuracy in predicting PDE-driven sequences compared to existing approaches, as well as competitive performance on Moving MNIST and BAIR benchmarks, revealing limitations of some compressed latent-space models.
    \item We conduct interpretability experiments that identify network regions encoding measurable physical dynamics. Further, we probe internal model representations to accurately estimate out-of-distribution simulation parameters, highlighting a learned encoding of underlying physics not directly extractable in pixel space.
\end{itemize}

\section{Related Works}\label{related_works}
We briefly summarise recent progress in video generative models, transformer-based approaches for video prediction, and video prediction models of physical systems specifically. 

\subsection{Video Models}
Research into unsupervised video models explores feature learning from images and videos \citep{Oord2018RepresentationLW, Donahue2019LargeSA, ranzato2014video, Chen2020GenerativePF, Oord2017NeuralDR}, as well as exhibiting improved downstream task performance from unsupervised pretraining \citep{Chen2020GenerativePF, wu2022nuwa, hong2023cogvideo}. A new category of image reconstruction loss functions \emph{DeePSiM} \citep{Dosovitskiy2016GeneratingIW} calculates image differences based on features extracted from pretrained image models, helping to mitigate smoothing artefacts observed when using image space distance metrics. Existing approaches utilising convolutional architectures include \citep {Amersfoort2017TransformationBasedMO, wang2020probabilistic}.

Studies adapting transformer models for image and video classification \citep{xie2021segformer, dosovitskiy2021an, bertasius2021space} tend to involve a patch-based framework by which input images are split into image patches and linearly embedded by a sequence of higher-dimensional 1D vector representations, as is typical for input to a transformer model. Feature learning and classification is performed exclusively using the transformer architecture, as such the patch-wise processing helps control parameter efficiency and locality. Purely transformer-based approaches distinguish themselves from CNN-based feature learning primarily in how they handle spatial and temporal locality. Unlike CNNs, which directly capture locality through convolutional layers, locality as a structural prior is restricted to intra-patch locality captured by feed-forward fully-connected layers. Predicting high-resolution future frames by leveraging pretrained image generators and a latent video prediction architecture \citep{seo2022harp} has been shown to significantly reduce training costs and improve prediction quality for various datasets.

\subsection{Autoregressive Video Models}
Autoregressive video prediction models have emerged as a prominent approach for forecasting future frames in video sequences, with notable examples including \citep{Weissenborn2020Scaling, Gao_2022_CVPR}. Models for autoregressive video prediction can be broadly defined as pixel-based models \citep{Chen2020GenerativePF, 10.5555/3045390.3045575, Gao_2022_CVPR}, or compressed latent models \citep{Rakhimov2021LatentVT,yu2023magvit}. Transformer-based models for video prediction often employ an \emph{encoder-predictor-decoder} structure, whereby a convolutional encoder model compresses the input image representation, with the transformer component processing the extracted features as a causal predictor model \citep{yan2021videogpt, Rakhimov2021LatentVT, seo2022harp}. An example of this is VideoGPT \citep{yan2021videogpt}, which uses a Vector-Quantised Variational Autoencoder (VQ-VAE) encoding strategy with a transformer model processing the compressed discrete latent representations as a causal frame predictor. SimVP \citep{Gao_2022_CVPR} uses a Vision Transformer (ViT) \citep{dosovitskiy2021an} as a predictor model, with CNN encoder/decoders for spatial feature extraction. IAM4VP \citep{seo2023implicitstackedautoregressivemodel} incorporates a stacked autoregressive approach, demonstrating improved performance in preserving temporal coherence and reducing error accumulation over long prediction horizons.

A significant shortcoming of these approaches remains the inherent propagation of errors accumulated at each prediction time-step, resulting in difficulties modelling longer sequences due to out-of-distribution predictions diverging from ground truth \citep{oprea2020review}. This is evident when observing the physical accuracy of predictions over time, and hence why we focus on evaluating model performance on predictions involving physical dynamics.

\subsection{Dynamic Simulation Modelling}
Introductory work exploring physically accurate video prediction models from unsupervised training \citep{10.5555/3157096.3157104} involves pixel-motion estimation and action-conditioning to generate future image frames. PhyDNet \citep{9156800} attempts to separate the modelling of videos of physical dynamics governed by Partial Differential Equations (PDE), from unknown residual information (e.g., texture, high-frequency details) by learning a semantic latent representation of the underlying PDE physics separate to variable pixel-based image information. The `Physics 101' dataset \citep{Wu2016Physics1L} is introduced to study physical properties of dynamic objects in video sequences, together with a model designed to explicitly encode physical laws via supervised parameter estimation. We leverage a set of physics-based video prediction datasets, where each dataset is associated with a parameter estimation task \citep{winterbottom2024powernextframepredictionlearning}. The DINo model \citep{yin2023continuous} introduces a data-driven approach for PDE forecasting that operates with continuous-time dynamics and spatially continuous functions, allowing learning from sparse and irregular data and generalise across different grids or resolutions. Other works exploring dynamic system modelling include \citep{10.5555/3305890.3306035, kolter2019learning, de2018deep}. A Fourier Neural Operator \citep{li2021fourier} is used to directly learn solutions to families of partial differential equations (PDEs) with high efficiency and accuracy, outperforming  previous learning-based PDE solvers. The DyAd model \citep{wang2022meta} employs meta-learning to improve generalisation in deep learning models for dynamics forecasting across varied domains, by partitioning them into distinct tasks.

Our approach distinguishes itself from the above due to a focus on reducing the need for structural priors specific to modelling physical dynamics, and instead focuses on a simpler self-supervised approach for end-to-end training.

\section{Autoregressive Video Prediction}\label{task_definition}
This section formalises the task definition for video prediction. Consider a sequence \( V \) consisting of image frames representing time-steps of a video. Let \(\mathcal{X} = \{x_t \mid t = 1, \dots, T\}\) be an input sequence comprising the first \( T \) time-steps of $V$, and \(\mathcal{Y} = \{y_{t'} \mid t' = T+1, \dots, T+T'\}\) be a target sequence consisting of the subsequent $T'$ frames of $V$, where each \( x_t \in \mathbb{R}^{C \times H \times W} \) and \( y_{t'} \in \mathbb{R}^{C \times H \times W} \) represents a $C$-channel image of height \( H \) and width \( W \). Given a video predictive model \(\mathcal{F} \)  parameterised by \( \theta \), \(\mathcal{F} \) is tasked with mapping the input sequence \( \mathcal{X} \) to the sequence of future frames \( \mathcal{Y} \). This can be performed in an autoregressive manner as follows:
\begin{equation}
\label{eq1}
\hat{y}_{t'} = \mathcal{F}(x_1, \dots, x_T, \hat{y}_{T+1}, \dots, \hat{y}_{t'-1}; \theta) ,
\end{equation}
\noindent where,
\begin{equation}
\label{eq2}
\hat{y}_{T+1} = \mathcal{F}(x_1, \dots, x_T; \theta) ,
\end{equation}
\noindent such that $y_{T+1}$ represents the first predicted frame, directly dependent on the input sequence. Each subsequent frame prediction $\hat{y}_{t'}$ up to time-step \( T+T' \) depends on all previous predictions \( \{y_{T+1}, y_{T+2}, \dots, y_{T+T'}\} \) and the original sequence \( \mathcal{X} \). The above describes the inference process of a model \(\mathcal{F} \) performing autoregressive video prediction given a input sequence \(\mathcal{X} \). 

Considering the learning objective of the autoregressive video prediction model described above, the training goal for \( \mathcal{F} \) is to minimise the average reconstruction loss between model predicted frames \( \hat{y}_{t'} \) and ground truth target frames \( y_{t'} \) for each time-step of the input sequence. During training, the input sequence \(\mathcal{X}\) consists of the entire video sequence \( V \) minus the final frame; \( \mathcal{X} = \{x_t \mid t = 1, \dots, T-1\} \), with targets \(\mathcal{Y}\) consisting of \( V \) minus the first frame; \( \mathcal{Y} = \{v_{t} \mid t = 2, \dots, T\} \), such that predictions at each time-step are conditioned only on ground truth frames and not model outputs. Our learning objective per video sequence \(\mathcal{X}\) can therefore be expressed as follows:
\begin{equation}
\label{eq3}
\min_{\theta} \sum_{t=1}^{T} \mathcal{L}\left(y_{t+1}, \mathcal{F}(x_1, \dots, x_{t}; \theta)\right) ,
\end{equation}
\noindent where \(\mathcal{L}\) represents the loss function.


\begin{figure*}[t!]
    \centering
    \subfloat[][]{\includegraphics[width=0.845\textwidth]{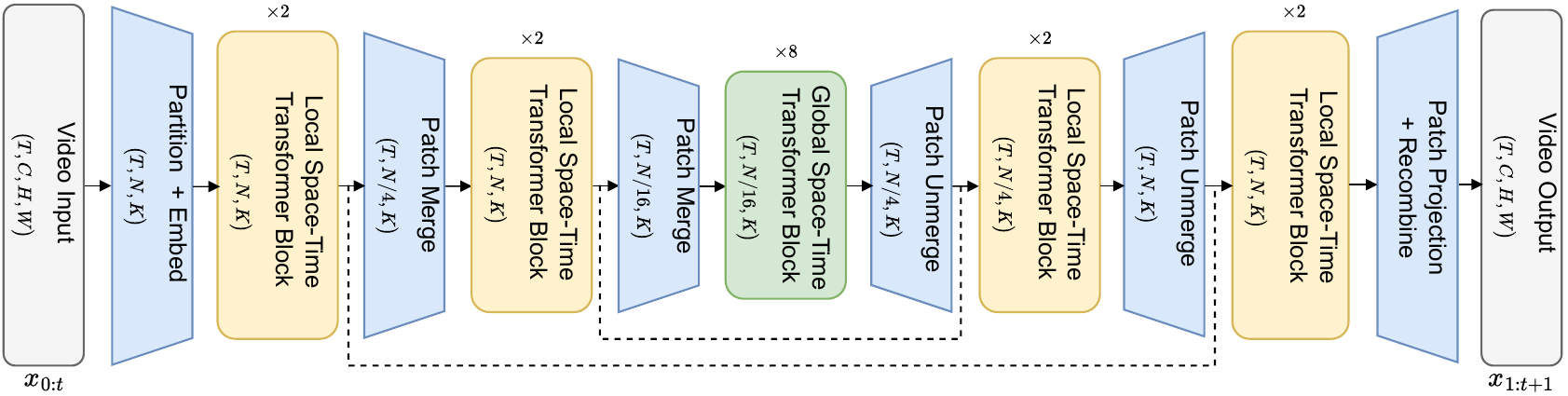}}
    \hfill
    \subfloat[][]{\includegraphics[width=0.13\textwidth]{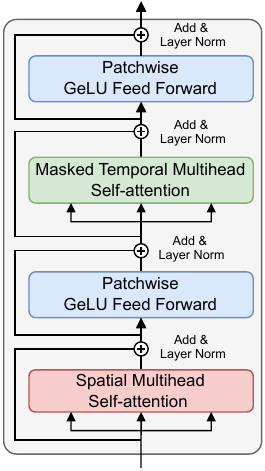}}
    \caption{\textbf{(a)} Overview of our PSViT model. \textbf{(b)} Space-time transformer layer. $T$ video frames are patched ($N$ patches), linearly embedded, then processed by local and global space-time transformer blocks. Local blocks use restricted attention windows with patch merge/unmerge (window size 2). Global blocks operate on a smaller effective image resolution. Skip connections between merge/unmerge operations preserve features. Input dimensions are annotated.}
    \label{fig2}
\end{figure*}

\section{Model Architecture}\label{model}
We build on the ViT \citep{dosovitskiy2021an} and TimesFormer \citep{bertasius2021space}, transformer-based models originally designed for image understanding and video understanding, respectively, and not generative modelling. In this section, we introduce our PSViT model, a pure-transformer backbone for testing different \emph{spatiotemporal self-attention schemes} for end-to-end unsupervised video prediction in \emph{continuous pixel-space}.

\subsection{Input Patch Processing}
The PSViT model (illustrated in Fig.~\ref{fig2}) takes as an input a video sequence $\mathcal{X} = \{x_t\}_{t=1}^T$, where each $x_t \in \mathbb{R}^{C \times H \times W}$ is a $C$-channel image frame of height $H$ and width $W$, sampled at time-step $t$ for $t = 1, 2, \ldots, T$. Each frame $x_t$ is partitioned into $S$ non-overlapping, equally-sized patches of dimension $P \times P$, with $H$ and $W$ being divisible by $P$ to ensure $S = \frac{HW}{P^2}$ patches that span the input image with no padding. Therefore, for each image frame $x_t$, we denote $x_{t, s} \in \mathbb{R}^{C \times P \times P}$ as the resulting image patch at index $s$ for time-step $t$, where $s = 1, 2, \ldots, S$ refers to the spatial location of the patch.

\subsection{Patch Embedding Representation}
Transformer models typically expect a sequence of 1D vector inputs. Following the partition of each frame into patches, each patch is further flattened into a 1D vector \( \bm{x}_{t, s} \in \mathbb{R}^{C  P^2} \) and linearly embedded into a higher-dimensional embedding space $D$ via a parameterised embedding matrix \( \bm{E} \in \mathbb{R}^{(C P^2) \times D} \):
\begin{equation}
\bm{z}_{t, s}^0 = \bm{E} \bm{x}_{t, s} ,
\label{eq5}
\end{equation}
\noindent where $\bm{z}_{t, s}^0 \in \mathbb{R}^D$ is the patch embedding vector of dimension $D$. For spatial and temporal contextual awareness, individual learnable positional encodings for each are added to the patch embedding as follows:
\begin{equation}
\bm{z}_{t, s} = \bm{z}_{t, s}^0 + \bm{e}_t^{\text{time}} + \bm{e}_s^{\text{space}} ,
\label{eq6}
\end{equation}
\noindent where $\bm{e}_t^{time} \in \mathbb{R}^D$ and $\bm{e}_s^{space} \in \mathbb{R}^D$ are parameterised encodings for time-steps $t = 1, 2, \ldots, T$ and spatial positions $s = 1, 2, \ldots, S$, respectively. The resulting sequence of patch embedding vectors $Z = \{\bm{z}_{t,s} \mid t=1,\ldots,T; s=1,\ldots,S\}$ serves as input to the transformer model, with the embedding process enabling the joint learning of both spatial and temporal information necessary for modelling video sequences.

\subsection{Spatiotemporal Attention Strategies}
Essential to our autoregressive video prediction model is the modification of \emph{multi-head self-attention} (hereafter denoted \emph{self-attention}) layers \citep{vaswani2017attention} to incorporate both intra-time-step spatial relationships, and inter-time-step causal temporal relationships across successive frames. We use a patch-based model similar to ViT, therefore we can view any spatial and temporal operations as capturing patch-wise spatial and temporal relationships. Dividing the input frame into patches is an effective method to avoid the quadratic growth in complexity associated with increasing image resolutions. We experiment with a range of spatiotemporal self-attention variations, building on those found in \citep{bertasius2021space}. Fig.~\ref{fig3} illustrates these spatiotemporal self-attention layouts. A key consideration when adapting these layouts for autoregressive video prediction is to enforce temporal causality via masking of patches from future time-steps during training.

Through preliminary testing, we find that combining spatial and temporal information in a single (global \emph{or} local) self-attention operation (\emph{Joint-ST} configuration \citep{bertasius2021space}) performs considerably worse when compared to isolating spatial and temporal information to separately parameterised self-attention layers. We further separate spatial and temporal self-attention operations by an additional patch-wise GeLU \cite{hendrycks2016gaussian} Feed-Forward Network (FFN), further explained in Section~\ref{output_layer}.

\subsection{Spatial Attention} 
Each spatial attention layer performs self-attention over image patch inputs independently for each time-step. This approach allows the layer to learn both local and global patchwise relationships, agnostic of temporal information. We follow a similar procedure for \emph{query}, \emph{key}, \emph{value} self-attention as described in \citep{vaswani2017attention, dosovitskiy2021an, bertasius2021space}. Given patch inputs $\bm{z}_{t,s}$, we have:
\begin{equation}
\label{eq7}
[\bm{k}_{t,s}, \bm{q}_{t,s}, \bm{v}_{t,s}] = \bm{z}_{t,s} \bm{U}_{kqv} \in \mathbb{R}^{3D_{\text{head}}} ,
\end{equation}
\noindent where key, query, and value vectors are $\bm{k} \in \mathbb{R}^{D_{\text{head}}}$, $\bm{q} \in \mathbb{R}^{D_{\text{head}}}$, and $\bm{v} \in \mathbb{R}^{D_{\text{head}}}$, respectively, $D_{\text{head}}$ is the head dimension for multiheaded attention, and $\bm{U}_{kqv} \in \mathbb{R}^{D \times 3D_{\text{head}}}$ is a linear projection matrix. Note that $\bm{U}_{kqv}$ is shared across time-steps. We then calculate spatial attention as follows:
\begin{equation}
\label{eq8}
\bm{z}^{\text{space}}_{t,s} = \operatorname{SM}\left(\frac{\bm{q}_{t,s} \cdot \begin{bmatrix} \bm{k}_{t,1}^\top & \bm{k}_{t,2}^\top & \cdots & \bm{k}_{t,S}^\top \end{bmatrix}}{\sqrt{D_{\text{head}}}}\right) \cdot \begin{bmatrix} \bm{v}_{t,1} \\ \bm{v}_{t,2} \\ \vdots \\ \bm{v}_{t,S} \end{bmatrix} ,
\end{equation}
\noindent where $\operatorname{SM}$ is the softmax operator, and $\bm{z}^{\text{space}}_{t,s}$ is a single head spatial attention output for spatial patch $s$ and time-step $t$. For Local-Space attention (shown in Fig.~\ref{fig3}), spatial patch indices not included in the computation are masked. We follow typical multi-headed self-attention practices for unifying attention heads and applying layer normalisation.

\subsection{Causal Temporal Attention}
To enable generative autoregressive behaviour, temporal self-attention is causally masked, allowing each patch to attend only to its own past and current time-steps across different frames, preventing information leakage from the future. Similar to spatial attention, for each spatial patch we apply temporal attention across time-steps, but limited to corresponding spatial patch positions. Due to better observed performance in preliminary  testing, we differ from existing approaches and learn separate temporal query, key, and value projections calculated from the output of the preceding spatial attention layer. 

\subsection{Layer Outputs}\label{output_layer}
The output for each patch at each intermediate layer $\ell$ for $L$ layers is as follows: 
\begin{equation}
\label{eq11}
\bm{z}_{t,s}^\ell = \operatorname{FFN'}(\bm{z}^{\text{time}}_{t,s}(\operatorname{FFN}(\bm{z}^{\text{space}}_{t,s}))) ,
\end{equation}
\noindent where $\operatorname{FFN}$ and $\operatorname{FFN'}$ are patchwise non-linear FFN components with layer normalisation and residual connections, typical of those found on the output of the original transformer layers. We note the following modifications to existing approaches: we further separate the computation of spatial and temporal attention via an FFN layer between spatial and temporal attention operations, and perform spatial attention before temporal attention due to consistently better results. Results are reported for ablations of this setup. The final layer $\operatorname{FFN'}$ output is down-projected to $C \times P \times P$ and reshaped to the original image patch dimensions, with patches at each time-step being reconstructed to form an output image $\hat{y}_{t'} \in \mathbb{R}^{C \times H \times W}$ as illustrated in Fig.~\hyperref[fig2]{2(a)}.

\subsection{U-net Style Adaptation}
We further adapt the typical transformer backbone architecture to progressively process patches at smaller resolutions, similar to that of U-net style models, such that adjacent input patches are linearly merged to reduce the effective image resolution at each block on the encoding side. This design choice allows us to train end-to-end models efficiently without resorting to pretrained encoder/decoder models. As illustrated in Fig.~\hyperref[fig2]{2(a)}, partitioning into patches is followed by a series of \textit{local} space-time transformed blocks separated by patch merge operations, with a \textit{global} space-time transformer block operating on the reduced resolution patch-based representations. This has the benefit of capturing multiscale spatial features, while reducing the computational complexity of the global self-attention operations due to a smaller number of image patches. The reverse process is applied on the decoding side of the model to produce a predicted next image frame for each input frame.

\begin{figure}[t!]
    \begin{center}
        \includegraphics[width=0.49\textwidth]{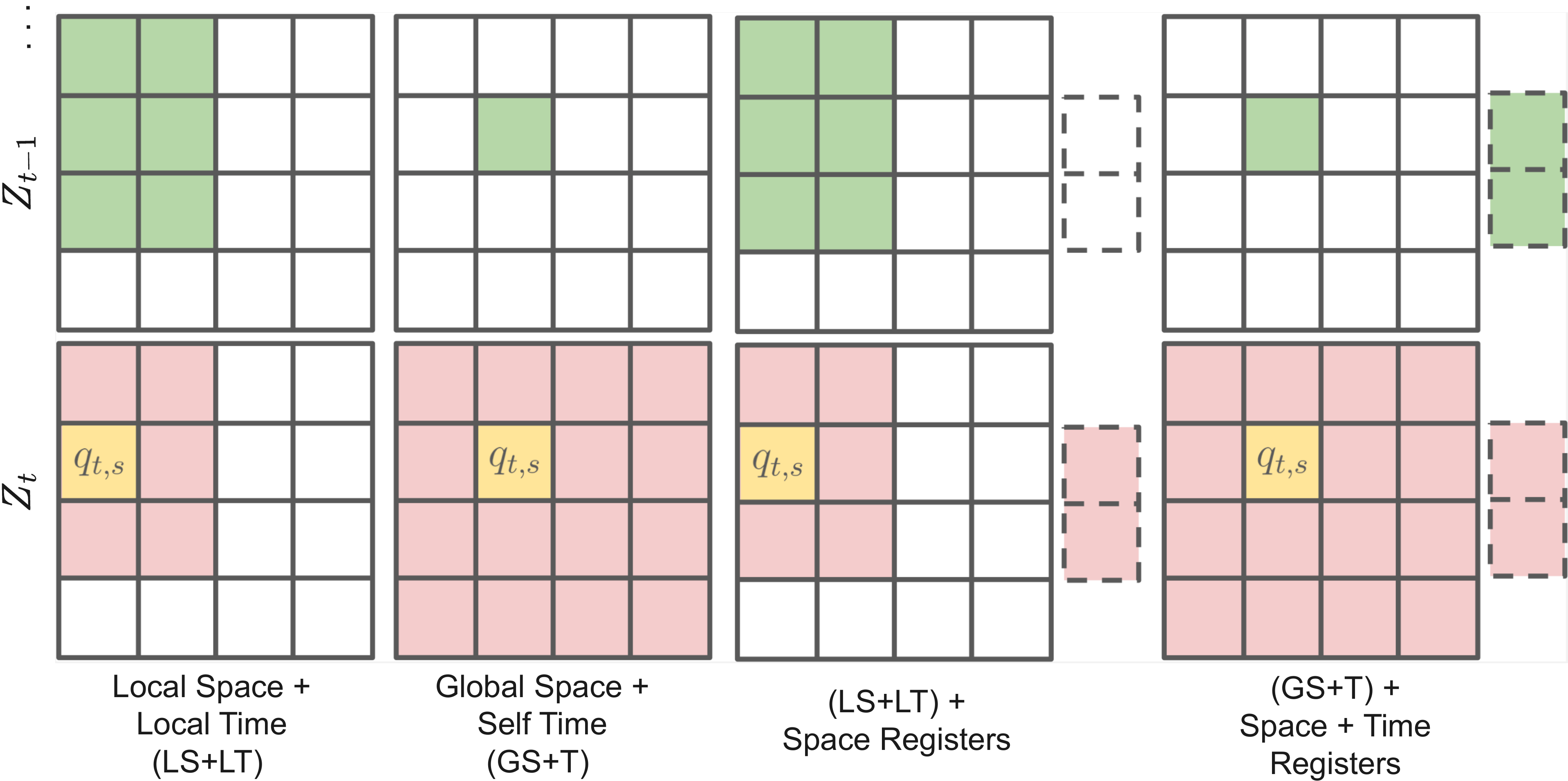} \\
    \end{center}
    \caption{Spatiotemporal self-attention masking strategies for 16-patch video frames, including register tokens (dashed) for sequence-level information. For a query patch $q_{t,s}$ (yellow) at time-step $t$, patch $s$: red patches are for spatial attention, green for temporal attention, and white are unused/masked. Previous frames (processed similarly) and future causally-masked frames are omitted.}
    \label{fig3}
\end{figure}

\subsection{Register Tokens}
We experiment with the addition of a set of register tokens \citep{darcet2024vision} added to the sequence embedded image patches at each time-step as a possible encoding mechanism for sequence-level information and PDE dynamics. $N$ learnable register tokens $\bm{r}_n \in \mathbb{R}^D, \text{ for } n = 1, 2, \ldots, N$ are appended to the set of $S$ patches for each time-step $t$ after the positional encodings step (Eq.~\eqref{eq6}), such that our input to the transformer becomes:
\begin{equation}
\label{eq12}
Z = [\bm{z}_{t, 1},\ldots, \bm{z}_{t,S}, \bm{r}_{t,1},\ldots, \bm{r}_{t,N}] \text{ for } t = 1,\ldots, T ,
\end{equation}
\noindent These tokens are discarded before recombining image patches to generate an output image. The intuition behind incorporating these tokens is such that they act as dedicated “memory” slots that aggregate non-local, sequence-level signals—such as global boundary conditions or slowly varying PDE modes, which can be accessed by each spatial patch.

\section{Experimental Setup}\label{methodology}
This section covers the datasets used for unsupervised video prediction training, outlines our experimental setups for training and evaluation, and covers the different model configurations we use.

\subsection{Datasets}
Our work focuses on unsupervised prediction of physics-based simulations involving objects moving and interacting according to well-defined physical laws. Specifically, we employ a set of physics-based \emph{simulation} datasets introduced in \citep{winterbottom2024powernextframepredictionlearning}. These datasets offer a controlled and visually-simple set of dynamic PDE simulations (calculated using the Runge-Kutta method \citep{butcher1996history}), useful for isolating physical accuracy of frame predictions. Each sequence per dataset is generated under unique initial conditions and simulation parameters, ensuring no parameter contamination between train, validation, and test splits. We use the following datasets: \emph{Moon} - an orbital dynamics simulation involving a static body and an orbiting moon, \emph{Pendulum} - a swinging pendulum under gravity, \emph{Roller} - a rolling ball on a curved surface acting under gravity, and \emph{3D Balls} - a 3D environment containing elastic collisions between moving balls and the environment walls. A full description of each dataset and the associated simulation parameters can be found in Appendix \ref{App:1}.

Each simulation dataset contains 3-channel image frame sequences at a resolution of 128
× 128 pixels. We split into training, evaluation, and testing subsets following an 80/10/10 split, ensuring no overlapping initial conditions or parameters across the splits. In addition to the simulation datasets above, we experiment with video prediction on the CLEVRER \citep{Yang_2019_CVPR} colliding objects dataset, and the Fluid simulation from DPI-Net \citep{li2019learning}. Additionally, we benchmark our model on two common video prediction benchmark datasets, namely Moving MNIST \citep{10.5555/3045118.3045209}, and BAIR robot pushing datasets \citep{ebert2017self}. For the CLEVRER, Fluid, Moving MNIST, and BAIR datasets, we use predefined training and testing splits. Specifically, for CLEVRER and Fluid, where the original resolution exceeds 128 $\times$ 128 pixels, we apply a central square crop and downsample to 128 $\times$ 128. 

\subsection{Object Divergence Metrics}
For each simulation dataset as well as the CLEVRER and Fluid dataset, we perform evaluation using an object divergence metric, whereby the centroid positions of objects contained in the observed scene are compared between predicted and ground truth sequences. This allows us to have an image quality invariant evaluation of object positional prediction over time. For each predicted time-step during inference, the centroid position of each object in the image frame is compared to its ground truth position via 2D euclidean pixel distance. A rolling average pixel divergence up to $t$ time-steps is taken as the \emph{Divergence} score up to $t$, with lower indicating closer prediction to ground truth. We normalise scores assuming a 128$\times$128 resolution to allow for fair comparison. For our main results, we evaluate using $t=50$. It is worth noting that the CLEVRER dataset involves frames in which moving objects appear from outside of view, and therefore are not predictable and not considered for object divergence scores. Additionally, the Fluid dataset contains groups of many particles which collide - we consider the centroid of each group for tracking purposes.

\begin{table}[ht!]
    \centering
    \caption{\noindent Model Configurations.}
    \label{tab:1}
    \resizebox{0.48\textwidth}{!}{%
    \begin{tabular}{l l c c c c c c}
        \toprule
        Model && Global ST Layers & Heads & Model dim. $D$ & Head dim. &  FFN dim. & Param. \\
        \midrule
        PSViT small && 8 & 12 & 512 & 64 & 2048 & 49M \\
        PSViT medium && 12 & 12 & 512 & 128 & 2048 & 84M \\
        \bottomrule
    \end{tabular}}
\vspace{-5pt}
\end{table}

\subsection{Training Setup}
For all experiments and model configurations (Table \ref{tab:2}), training is performed using the Adam optimiser \citep{KingBa15}, a weight decay factor of $1e^{-4}$, and a batch size of 32. These parameters were selected following a thorough evaluation of hyperparameters across various datasets and models. Each model is trained using the Structural Similarity Index Measure (SSIM) \citep{1284395} loss function (experiments using other loss functions were conducted, with SSIM generalising best to all datasets), with input pixel values normalised to the range $[0, 1]$. We train each model for a maximum of 2000 epochs. For simplicity, we do not include any additional data augmentation techniques. All models are trained on an NVIDIA A100 GPU. Details on model training and inference costs can be found in the Appendix.

\subsection{Comparison to Existing Approaches}
\label{subsec:baselines}

We compare our method against state-of-the-art video prediction approaches (model parameter sizes in brackets). For latent-space baselines, we use: \textit{a)} \textit{MAGVIT} (300M) \citep{yu2023magvit}, which tokenises clips with a 3-D VQ-VAE and reconstructs them non-autoregressively via masked-token prediction with a transformer; \textit{b)} \textit{Latent Diffusion Transformer} Open-Sora (1B) \citep{zheng2024open}, whose transformer denoiser learns spatiotemporal dynamics directly on compressed video latents; and \textit{c)} \textit{CV-VAE} (160M) \citep{zhao2024cvvae}, a continuous 3-D video VAE made compatible with pre-trained image VAEs for efficient spatiotemporal latent modelling. Additionally, we benchmark against \textit{SimVP} (33M) \citep{Gao_2022_CVPR}, a convolutional encoder–decoder that autoregressively forecasts pixels without recurrence or attention. We follow the original authors' training protocols for each baseline wherever possible. These four baselines span key design paradigms of modern video prediction: latent vs.\ pixel space, and non-autoregressive, autoregressive, and diffusion transformer architectures.

\section{Results}\label{results}
This section evaluates our PSViT model for autoregressive video prediction. We begin with ablation studies to determine optimal PSViT architectural choices, followed by broader performance evaluations against existing approaches and further analyses. All results are from held-out test sets.

\subsection{Ablation Studies on Model Configuration}
We conduct ablation studies to determine optimal positional encodings, spatiotemporal self-attention strategies, and patch sizes $P$. These results (Table~\ref{tab:2}, top panels) guided our final architecture selection for subsequent comparisons.

\subsubsection{Positional Encodings}
A common approach is to use absolute positional embeddings for each token, \emph{e.g.,} in the form of a periodic sinusoidal function, or a Learnable Positional Encoding (LPE) via a parameterised embedding trained jointly with the model \citep{zhang2022opt}. Relative embeddings via positional lookup tables are used in \citep{raffel2020exploring}, and Rotary Positional Embeddings (RoPE) \citep{chowdhery2023palm} perform rotational operations on the query and key self-attention matrices, using angular values from absolute positions. We experimented with APE, RoPE, and LPE for both spatial and temporal positional encodings. Table~\ref{tab:2} shows LPE yielded the best average object divergence score, demonstrating a clear benefit. This suggests that allowing the model to learn the optimal way to encode positional information is advantageous for these tasks. Consequently, LPE is used for all PSViT configurations and results hereafter. 

\begin{figure}[t!]
    \centering
    \includegraphics[width=0.49\textwidth]{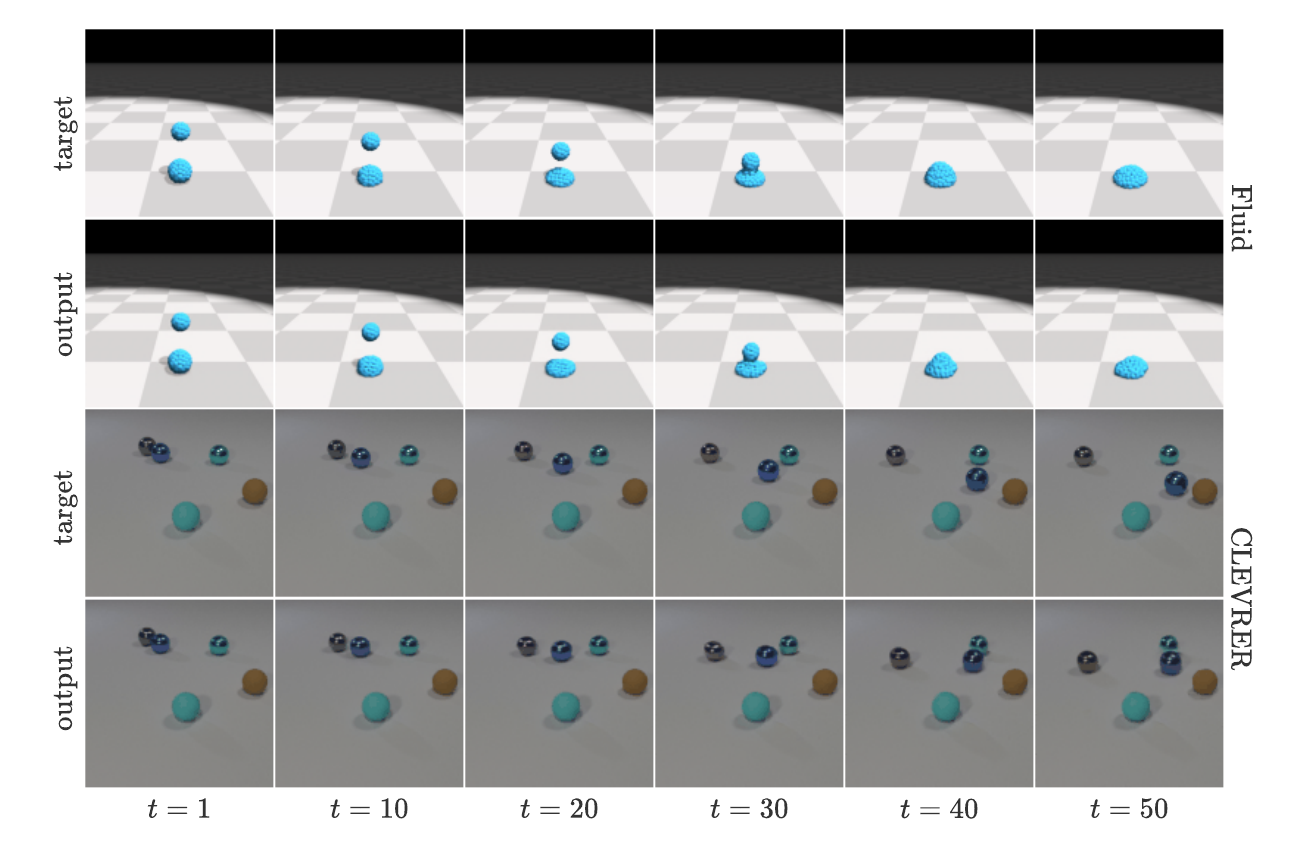} \\
    \caption{PSViT sample outputs vs. ground truth for Fluid and CLEVRER datasets, conditioned on 12 input frames (intermediate time-steps omitted).}
    \label{fig4}
\end{figure}

\begin{figure}[t!]
    \centering
    \includegraphics[width=0.49\textwidth]{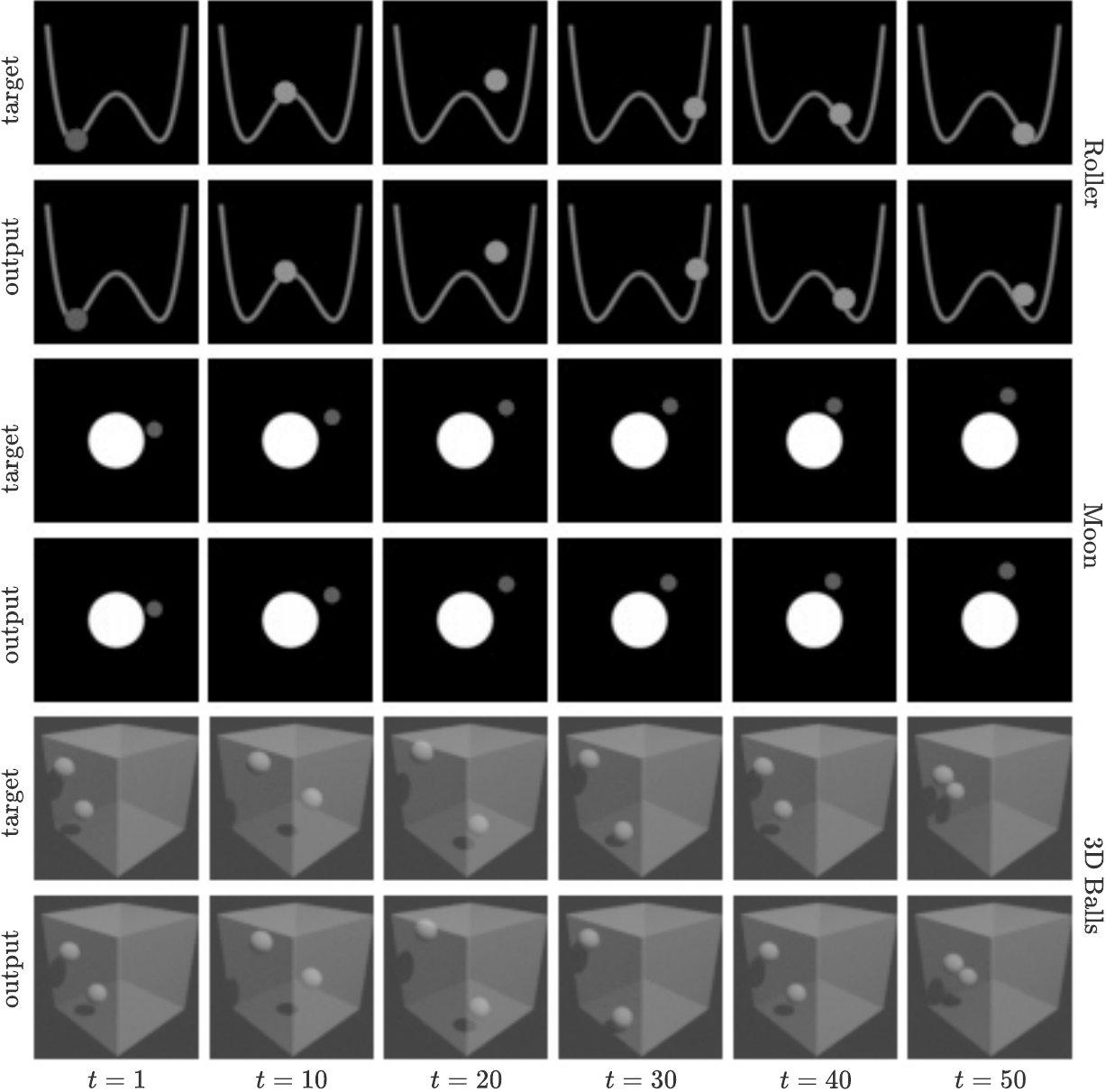} \\
    \caption{PSViT sample outputs (representative of median divergence performance) vs. ground truth for Roller, Moon, and 3D Balls datasets, conditioned on 12 input frames (intermediate time-steps omitted).}
    \label{fig5}
\end{figure}

\begin{table*}[t!]
    \centering    
    \caption{Video Prediction Results and Divergence Scores}
    \label{tab:2}
    \resizebox{0.85\textwidth}{!}{%
    \begin{tabular}{l l c c c c c c r}
        \toprule
        \\[-1\medskipamount]
         &&&& \multicolumn{2}{c}{\textbf{Divergence (\emph{$t=50$})} $\downarrow$} &&& \\
        \noalign{\smallskip}
        \cline{5-6}
        \noalign{\smallskip}
        \textbf{Approach} && \multicolumn{1}{c}{\textbf{Roller}} & \multicolumn{1}{c}{\textbf{Moon}} & \multicolumn{1}{c}{\textbf{Pendulum}} & \multicolumn{1}{c}{\textbf{3D Balls}} & \multicolumn{1}{c}{\textbf{Fluid}} & \multicolumn{1}{c}{\textbf{CLEVRER}} & \multicolumn{1}{r}{\textbf{Avg}} \\
        \midrule
        PSViT Attn. Strategy \\
        \hspace{1pt} - LS + LT                      && 1.52 & 2.20 & 2.08 & 5.83 & 2.54 & 6.01 & 3.36 \\
        \hspace{1pt} - GS + LT                      && 1.32 & 2.17 & 1.94 & 5.41 & 2.31 & 5.30 & 3.08 \\
        \hspace{1pt} - GS + T                       && 1.30 & 2.18 & 1.92 & 5.41 & 2.28 & 5.28 & \textbf{3.06} \\
        \midrule
        PSViT Pos Encoding \\
        \hspace{1pt} - APE                          && 1.41 & 2.18 & 2.03 & 5.57 & 2.34 & 5.71 & 3.21 \\
        \hspace{1pt} - RoPE                         && 1.30 & 2.16 & 1.97 & 5.37 & 2.24 & 5.41 & 3.08 \\
        \hspace{1pt} - LPE                          && 1.28 & 2.17 & 1.92 & 5.28 & 2.25 & 5.41 & \textbf{3.05} \\
        \midrule
        CV-VAE (160M) \citep{zhao2024cvvae}                && 1.18 & 2.10 & 1.98 & 5.35 & 2.88 & 5.29 & 3.13 \\
        Diffusion Transformer (1B) \citep{zheng2024open} && 1.20 & 2.14 & 2.06 & 5.42 & 2.95 & 5.35 & 3.19 \\
        MAGVIT (300M) \citep{yu2023magvit}                 && 1.17 & 2.20 & 2.12 & 6.02 & 2.54 & 5.40 & 3.24 \\
        SimVP (33M) \citep{Gao_2022_CVPR}                 && 1.19 & 2.12 & 2.04 & 5.39 & 2.93 & 5.32 & 3.16 \\
        PSViT small (49M)                                 && 1.28 & 2.17 & 1.92 & 5.28 & 2.25 & 5.41 & 3.05 \\
        PSViT medium (84M)                              && 1.18 & 2.06 & 1.84 & 5.10 & 2.04 & 5.18 & \textbf{2.90} \\
        \midrule
        &&&& \multicolumn{2}{c}{\textbf{SSIM $\uparrow$}} &&& \\
        \noalign{\smallskip}
        \cline{5-6}
        \noalign{\smallskip}
        MAGVIT \citep{yu2023magvit}                 && 0.9996 & 0.9996 & 0.9993 & 0.9943 & 0.9654 & 0.9803 & 0.9901 \\
        SimVP \citep{Gao_2022_CVPR}                 && 0.9996 & 0.9996 & 0.9994 & 0.9946 & 0.9674 & 0.9839 & 0.9908 \\
        PSViT Medium                                && 0.9998 & 0.9997 & 0.9997 & 0.9961 & 0.9752 & 0.9951 & \textbf{0.9943} \\
        \midrule
        &&&& \multicolumn{2}{c}{\textbf{PSNR $\uparrow$}} &&& \\
        \noalign{\smallskip}
        \cline{5-6}
        \noalign{\smallskip}
        MAGVIT \citep{yu2023magvit}                 && 59.70 & 56.56 & 55.59 & 50.99 & 47.03 & 50.76 & 53.44 \\
        SimVP \citep{Gao_2022_CVPR}                 && 58.42 & 56.60 & 55.36 & 50.53 & 48.23 & 50.21 & 53.23 \\
        PSViT Medium                                && 59.78 & 56.63 & 56.57 & 53.91 & 47.19 & 53.30 & \textbf{54.56} \\
        \bottomrule
    \end{tabular}}
\end{table*}

\subsubsection{Self-Attention Strategies and Patch Size}
The design of the spatiotemporal self-attention mechanism is central to how PSViT captures and models video dynamics. We experimented with three different spatiotemporal self-attention strategies (illustrated in Fig.~\ref{fig3}) for the middle space-time layers (the \textit{Global Space-Time Block} in Fig.~\hyperref[fig2]{2(a)}), namely local-space + local-time (LS+LT), global-space + self-time (GS+T), and global-space + local-time (GS+LT). While reducing the patch-wise receptive field at each layer, local self-attention operations benefit from linear complexity scaling relative to image resolution and choice of patch size $P$.

Table~\ref{tab:2} shows GS+T and GS+LT significantly outperformed LS+LT across all datasets. This clearly shows that reducing the global spatial receptive field is detrimental for these tasks. Restricting temporal attention to the same patch location across frames (the +T scheme in GS+T) results in performance comparable to attending to a local temporal window of patches (GS+LT), but with the added benefit of linear complexity with respect to sequence length. The difference in performance is most apparent for the more visually-complex 3D datasets, 3D Balls and CLEVRER. Given its strong performance and efficiency, the GS+T strategy was chosen for our standard PSViT model.

Regarding patch size, preliminary testing revealed that $P=8$ yields a better performance-efficiency trade-off than $P=16$ (which was used in the original ViT work \citep{dosovitskiy2021an}). Increasing the patch size beyond $P=8$ led to parameter inefficiency without a corresponding performance gain, while reducing it further (e.g., to $P=4$) resulted in degraded performance and prohibitively high memory consumption. Therefore, a patch size of $P=8$ is used for all PSViT models.

\subsection{Overall Video Prediction Performance}
Our full set of results for video prediction on the simulation datasets is presented in Table~\ref{tab:2}, with lower scores indicating better performance for object divergence, and higher indicating better for SSIM and PSNR. SSIM and PSNR scores are an average over the first 5 output frames, while divergence scores are taken after 50 output time-steps. All scores are test set averages, with best average performance for each comparison in bold.

Example model outputs representative of the median performance for object divergence are shown in Fig.~\ref{fig4} for the CLEVRER and Fluid datasets. We observe accurate object trajectories and interactions of multiple objects relative to ground truth frames up to $t=30$, after which we observe deviations to target object positions. Fig.~\ref{fig5} shows example outputs for the Roller, Moon, and 3D Balls datasets, where the same observation is made with accurate predictions up to $t=30$.

\begin{figure*}[t!]
    \centering
    \includegraphics[width=0.6\textwidth]{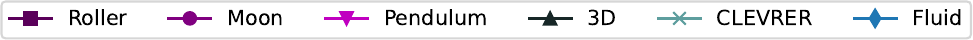}\\
    \subfloat[][\label{fig6a}]{\includegraphics[width=0.237\linewidth]{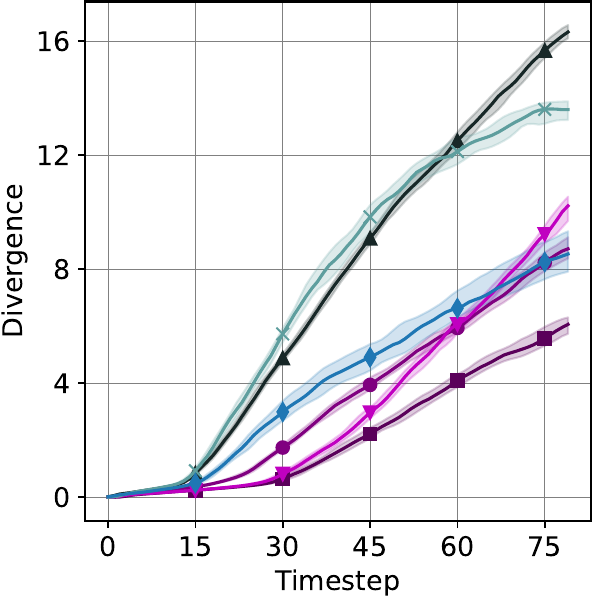}}
    \hfill
    \subfloat[][\label{fig6b}]{\includegraphics[width=0.241\linewidth]{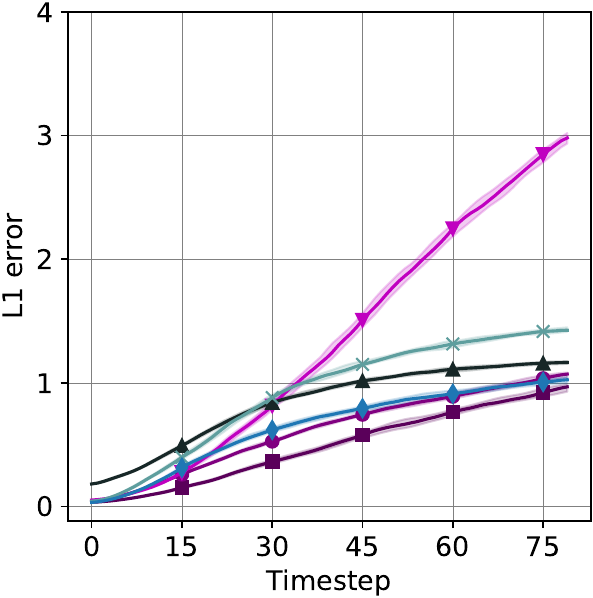}} 
    \hfill
    \subfloat[][\label{fig6c}]{\includegraphics[width=0.241\linewidth]{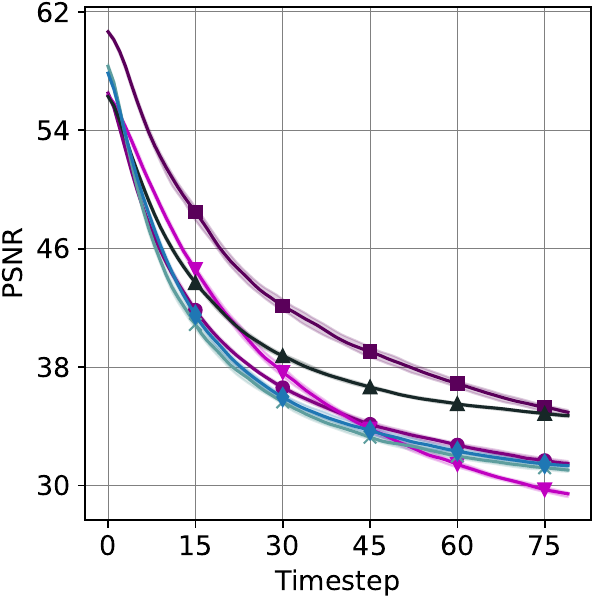}}
    \hfill
    \subfloat[][\label{fig6d}]{\includegraphics[width=0.237\linewidth]{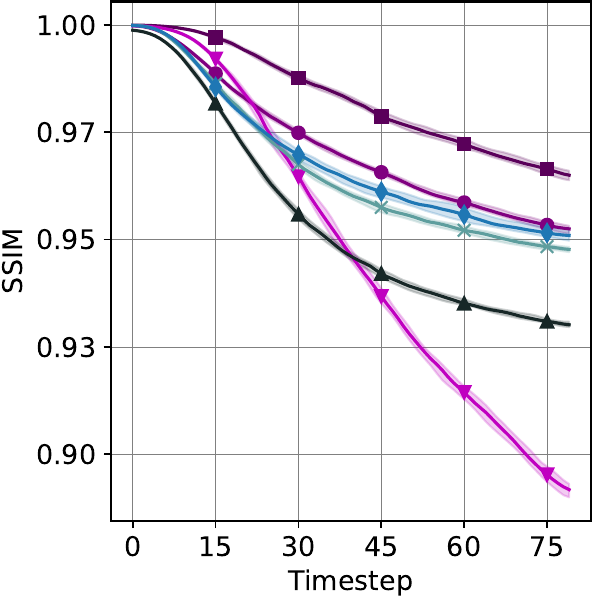}}
    \caption{Model performance metrics ((a) Object Divergence, (b) L1 Error, (c) PSNR, (d) SSIM) over 80 self-conditioned output time-steps, with 12 input frames. Object Divergence is the median rolling average over all test sequences.}
    \label{fig6}
\end{figure*}

When considering different model scales (PSViT-small and PSViT-medium, configurations in Table~\ref{tab:1}) and the inclusion of parameterised register tokens (as described in Section~\ref{model}), we observe a performance increase across all tasks (Table~\ref{tab:2}). Perhaps unsurprisingly, increasing model size has a bigger performance impact on the 3D datasets when compared to the 2D physics simulations, suggesting that the increased parameterisation is necessary for handling the increase in visual complexity rather than solely learning of PDE dynamics.

Fig.~\ref{fig6} compares performance at each output time-step across datasets for divergence, L1, SSIM, and PSNR. More generally, we note a large relative drop in performance for the 3D and CLEVRER datasets when measured by object divergence. Interestingly, this is a difference that is not captured as markedly using L1, SSIM, and PSNR metrics, suggesting that including our object divergence metric is an important factor in determining model performance regarding accuracy on the underlying physics. All tasks see low object divergence up to $t=20$, with the 2D simulation datasets slightly better to varying degrees. The divergence rate for the Roller, Moon, and Pendulum datasets correlates with the number of PDE variables; the Pendulum dataset, having five PDE variables, performs the worst, compared to the Roller dataset with two. Considering the relatively high performance of the Fluid dataset given the visual complexity, we attribute this to low variance between samples - object interactions typically occur near the centre of the image.

\begin{figure}[t!]
    \centering
    \includegraphics[width=0.48\textwidth]{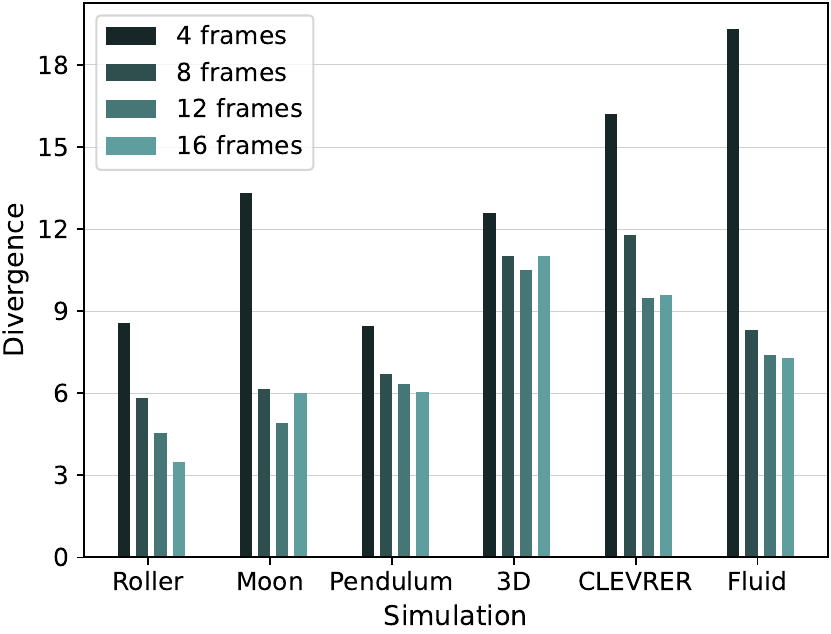}
    \caption{PSViT object divergence scores vs. input context size, averaged over test sets after 50 prediction time-steps.}
    \label{fig7}
\vspace{-10pt}
\end{figure} 

\begin{figure*}[t!]
    \centering
    \includegraphics[width=0.5\textwidth]{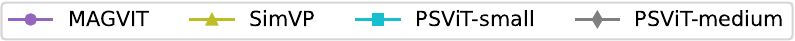}\\
    \subfloat[][]{\includegraphics[width=0.269\linewidth]{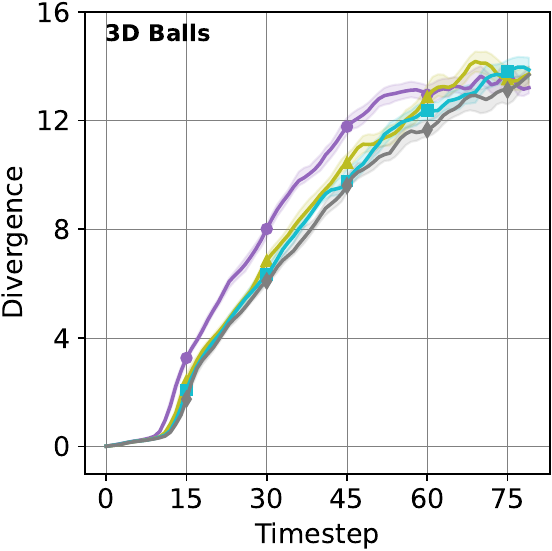}}
    \hfill
    \subfloat[][]{\includegraphics[width=0.232\linewidth]{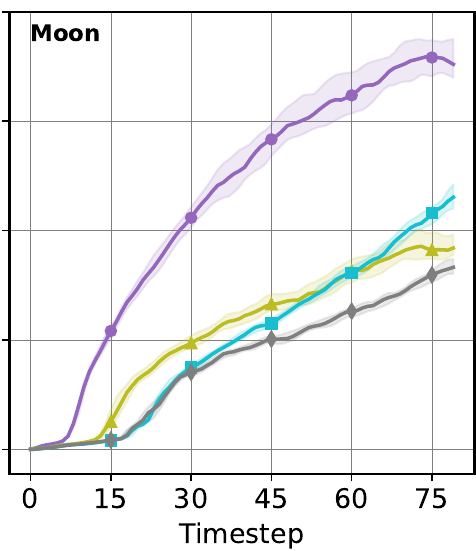}} 
    \hfill
    \subfloat[][]{\includegraphics[width=0.232\linewidth]{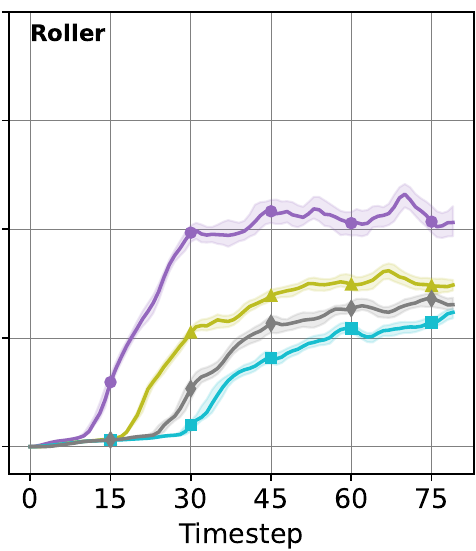}}
    \hfill
    \subfloat[][]{\includegraphics[width=0.232\linewidth]{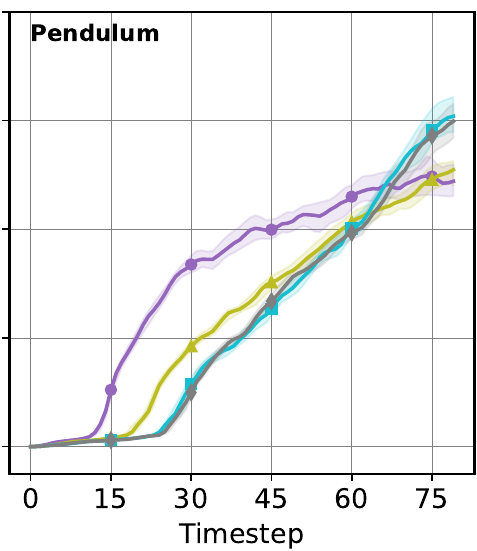}}

    \caption{Object divergence of our model (PSViT) vs. existing approaches (MAGVIT, SimVP) on (a) 3D Balls, (b) Moon, (c) Roller, and (d) Pendulum datasets.}
    \label{fig8}
\end{figure*}

\begin{figure}[ht!]
    \centering
    \includegraphics[width=0.48\textwidth]{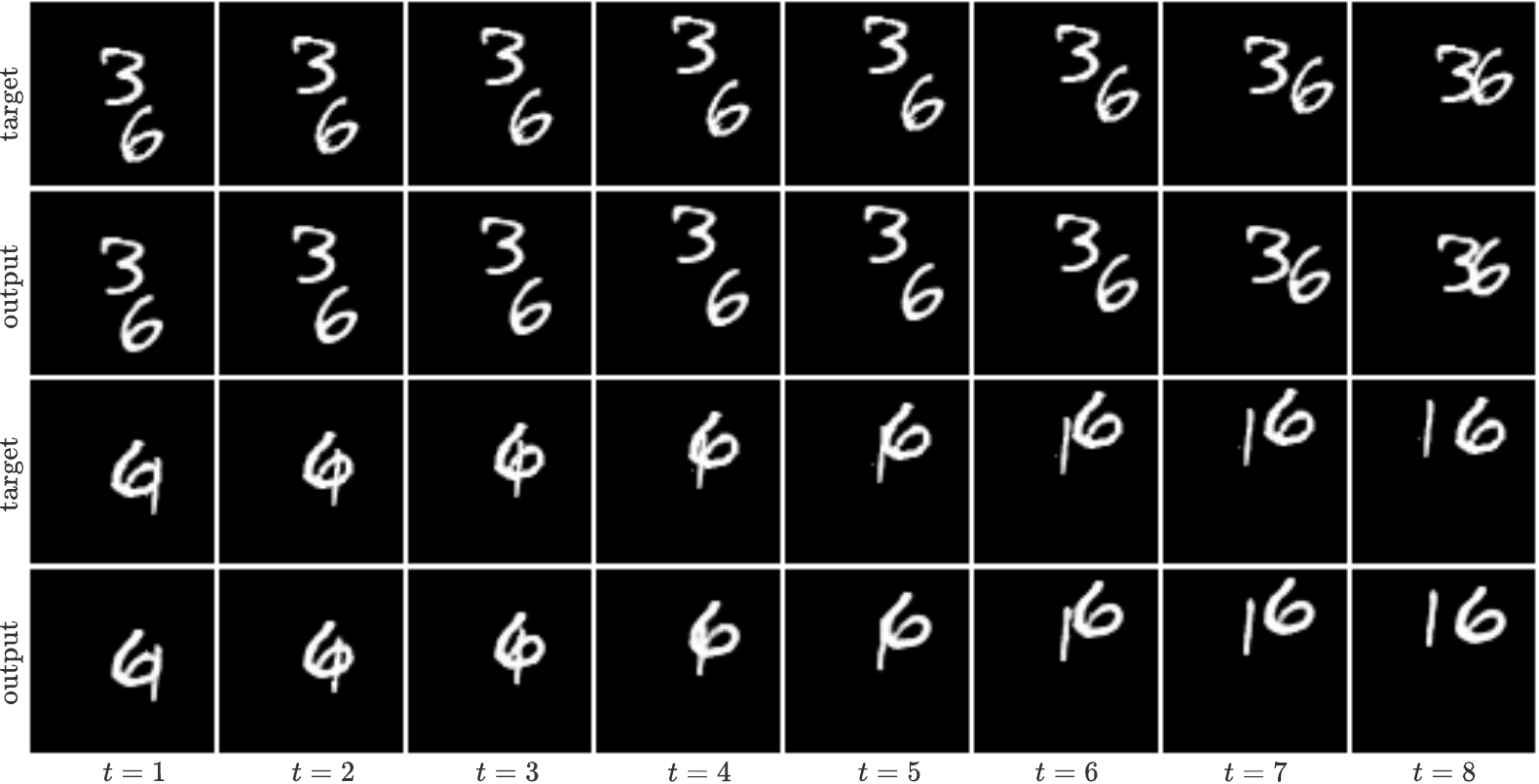}
    \caption{Randomly selected model predictions from the Moving MNIST dataset conditioned on 4 input frames using PSViT medium.}
    \label{mmnist}
\end{figure}

\begin{figure*}[t!]
    \centering
    \subfloat[][]{\includegraphics[width=0.27\textwidth]{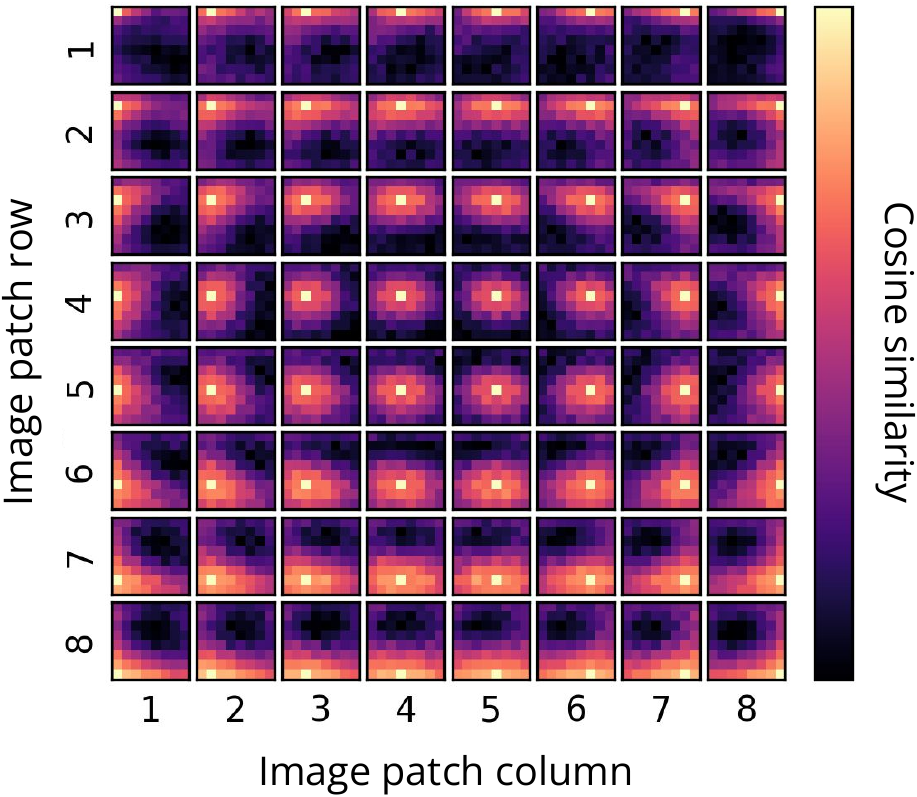}}
    \hfill
    \subfloat[][]{\includegraphics[width=0.10\textwidth]{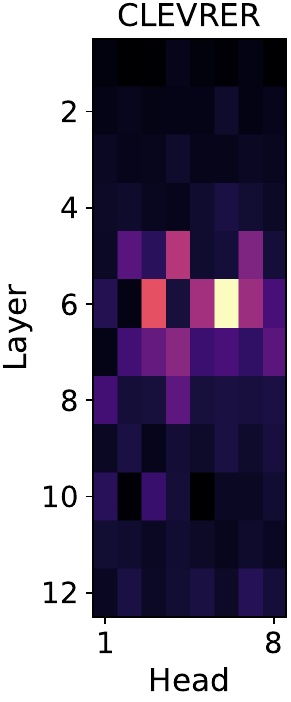}\includegraphics[width=0.074\textwidth]{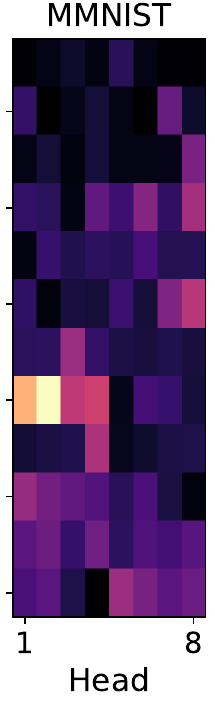}\includegraphics[width=0.117\textwidth]{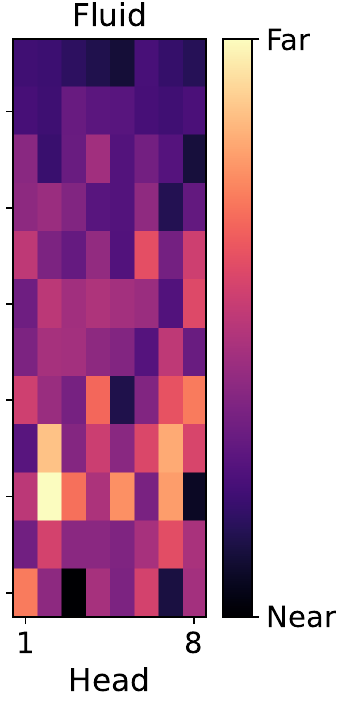}}
    \hfill
    \subfloat[][]{\includegraphics[width=0.38\textwidth]{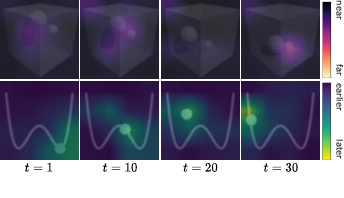}}
    \caption{\textbf{(a)} Visualisation of learned image-patch spatial positional encodings showing cosine similarity between patches.\textbf{(b)} Layer-wise spatial self-attention head weightings on CLEVRER, Moving MNIST, and Fluid datasets, with activations normalised by median patch distance scores, such that 0 is the query patch, and 1 is the most distant patch. \textbf{(c)} Attention activation heatmaps on example outputs for individual spatial (top, correlated with object interactions) and temporal (bottom, correlated with object velocity) self-attention heads.}
    \label{heatmaps}
\end{figure*}

\subsection{Comparing Input Context Sizes} 
We experiment training our model with different fixed input context sizes, shown in Fig.~\ref{fig7}. The impact of additional context frames differs significantly between tasks. Interestingly, we see that the maximum context size does not achieve best performance for three of the datasets studied, indicating a limitation on handling increased context length. We find that the number of context frames has a significant impact on the time-step at which divergence from ground truth begins to occur, but does not necessarily impact the rate of divergence past this point. The significant reduction in divergence between 4 and 8 frames of context for the Fluid dataset is expected due to the high number of objects being modelled.

\subsection{Comparison with Existing Approaches}
We compare our approach with recent state-of-the-art models covering different paradigms in video prediction, namely MAGVIT \citep{yu2023magvit} and SimVP \citep{Gao_2022_CVPR}, Latent Diffusion Transformer \citep{zheng2024open}, and CV-VAE \citep{zhao2024cvvae}. Fig.~\ref{fig8} compares object divergence performance between four models on the Roller, Moon, Pendulum, and 3D Balls datasets. Both sizes of our PSViT model performs significantly better across the first 50 time-steps in comparison to SimVP and MAGVIT, with the exception of 3D Balls dataset where all models have similar performances. We attribute this to the relative simple dynamics of the 3D Balls dataset, as all objects share a constant velocity across all sequences (only size and initial trajectory are varied). The most notable difference is the time-step at which divergence begins to increase compared to both existing approaches, with our approach significantly increasing the time horizon of accurate prediction on all datasets by up to 50\%. Averaged over all datasets, Table~\ref{tab:2} shows both scales of our approach perform favourably for Divergence scores against all comparisons, especially considering the parameter differences to models such as the Diffusion Transformer. 

\begin{table}[ht!]
    \centering
    \caption{Video prediction performance on BAIR and Moving MNIST}
    \label{tab:3}
    \resizebox{0.45\textwidth}{!}{%
    \begin{tabular}{l l c c}
        \toprule
        Approach && BAIR (FVD) & Moving MNIST (SSIM) \\
        \midrule
        SimVP \citep{Gao_2022_CVPR} && 67.1 & 0.948 \\ 
        MAGVIT \citep{yu2023magvit}  && 62.4 & 0.938 \\
        CV-VAE \citep{zhao2024cvvae} && 63.6 & 0.945 \\ 
        Diffusion Transformer \citep{zheng2024open} && \textbf{61.0} & 0.950 \\ 
        PSViT-medium (ours) && 64.1 & \textbf{0.963} \\
        \bottomrule              
    \end{tabular}}
\end{table}

In addition to the physical simulation datasets, we also compare our model on two common video prediction benchmark datasets, BAIR and Moving MNIST. Results are detailed in Table~\ref{tab:3}. Fréchet Video Distance (FVD) \citep{Unterthiner2019FVDAN} is used for the BAIR dataset. We observe competitive performance of our model on the BAIR probabilistic dataset - although the latent space approaches outperform both ours and SimVP. For the Moving MNIST dataset, the latent space approaches perform worse. This reaffirms that while latent models can produce high-quality video generation needed for stochastic datasets such as BAIR, they fall short in terms of physical coherence over time. 

\subsection{Qualitative Assessment} 
We perform a qualitative analysis of model outputs over time, and where they begin to fall short and diverge. Fig.~\ref{fig5} shows example model outputs from the visually-simpler physics simulation datasets. Our model is clearly capable of preserving the shape and colour of objects over time, regardless of divergence to ground truth. Prediction errors begin to appear at the later time-steps; for instance in the 3D Ball scenario shown there are clear inconsistencies between the shadows of the colliding objects. The same is true for the CLEVRER and Fluid examples both shown in Fig.~\ref{fig1}~and~Fig.~\ref{fig4}, where fine-detail errors increase in late output time-steps, although shadows and object reflections appear to be modelled successfully. Additionally, distortions in shape and incorrect modelling of object rotation following collisions is notable. The Moving MNIST examples shown in Fig.~\ref{mmnist} illustrate correct positioning of the characters, though finer details are smoothed out over time. Additional model output sequences are included in the Appendix.

\section{Discussion and Structural Analysis}\label{structural}
In the following section we investigate the representations learned by the patch-wise spatial and temporal self-attention layers. Each of these structures are responsible for attending to information across a range of temporal and spatial hierarchies. We assume the GS+T self-attention scheme throughout the remainder of this section. We investigate how these layers attend to information at different model depths, and what sequence-specific information can be extracted from these representations.

\subsection{Spatial and Temporal Reasoning}
Fig.~\hyperref[heatmaps]{10(a)} visualises the learned patch-wise spatial positioning encodings. We observe that the learned positioning encodings do not learn image-symmetrical representations for patches, such that patch-wise similarity does not necessarily scale linearly with patch distance. This can be observed in the corner and edge patch encodings, which show higher similarity to other edge patches than the middle patches, despite being further away. This is a learned distinction that a sinusoidal absolute encodings would not provide, suggesting a beneficial property for patches to explicitly encode edge-aware information directly into the embeddings. 

\begin{figure}[t!]
    \includegraphics[width=0.49\textwidth]{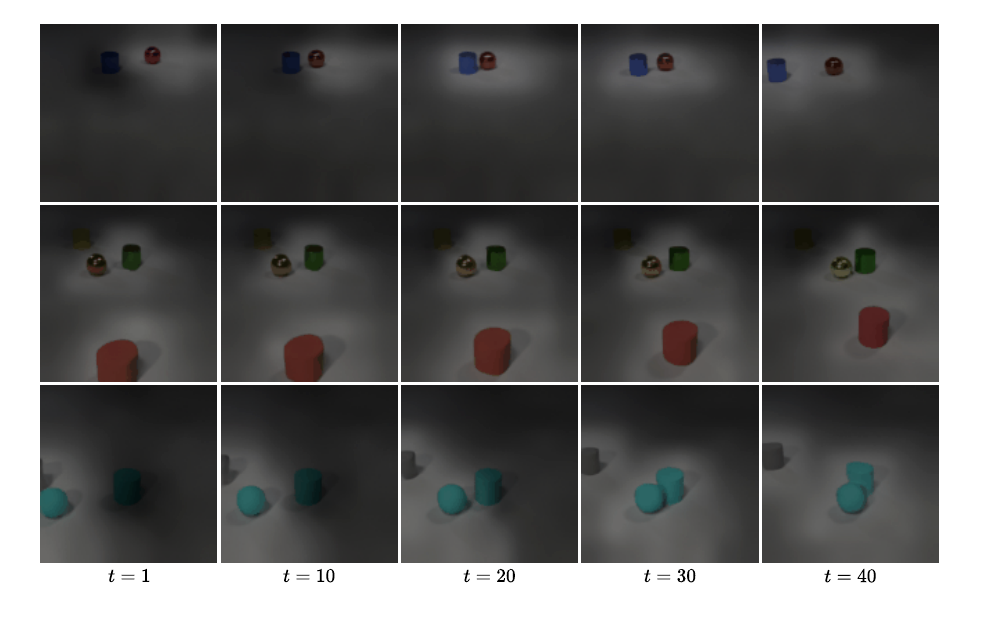}
    \caption{Spatial attention activation heatmaps from randomly selected model outputs on the CLEVRER dataset.}
    \label{fig11}
\end{figure}
 
Determining structural information encoding is an important step in model interpretability, where highlighting the information processed at different model depths can aid in understanding decision making and reasoning internal to the model. To this end, Fig.~\hyperref[heatmaps]{10(b)} shows a heatmap of the relative spatial self-attention distances at each layer and each attention head of the model. We show test-set averages for the CLEVRER, Moving MNIST, and Fluid datasets. We calculate this as the median distance over image patches for which each patch is attending to, averaged over all patches. Intuitively, this is the average attended receptive field per head. We observe a general trend of increased receptive field between the middle and final layers of the model. For instance, the CLEVRER example shows high spatial receptive field contained to the middle layers, with small receptive fields in the shallow and deep layers. A much broader distribution is observed for the Fluid dataset with an increased receptive field throughout the model, though the middle and deeper layers attend to distant patches.

\subsection{Attention Head Mechanisms}
In Fig.~\hyperref[heatmaps]{10(c)} we visualise the activation maps for a single spatial (top) and temporal (bottom) attention head on the 3D ball and Roller datasets, respectively. In the top row, we isolate an attention head correlating to object collisions, which can be seen by the attention activation map when the larger ball collides with the box boundary ($t=10$), and when the two balls approach and collide together ($t=20$, $t=30$), followed by reduced activation over these patches. Additionally, we identify a temporal self-attention head highly correlated to object velocity on the Roller dataset (bottom). We observe attention head activity on patches containing the moving object, where later time-steps are attended to with increased object velocity.

Fig \ref{fig11} shows spatial attention activations for register tokens averaged over all layers, clearly highlighting attention focus on moving objects.

\subsection{PDE Dynamics Information Probing}
In this section, we examine the internal representations of the space-time layers of our model to uncover the encoding and extractability of sequence-specific PDE parameters. These parameters are highly important for scene understanding and crucial for accurate object prediction over-time. We train linear probes on top of frozen intermediate representations extracted from each layer of our PSViT model in order to determine extractable information across internal model representations. Linear probing provides a straightforward approach for this, though we also experiment with non-linear estimators to explore what can be learned from internal representations.

Table~\ref{tab:4} reports our probing results for parameter estimation. We perform gravity estimation on the Roller and Pendulum datasets, and mass estimation on the Moon Orbit dataset. All probing results are an average over a held-out test set of 1,000 sequences. Baseline scores are taken from a randomly initialised equivalent model. Layers used are the global space-time layers. We observe that for all three tasks, middle-layer representations are capable of extracting higher performance on the regression tasks, with later layers performing the worst - presumably as these layers are closer to the output pixel regression layer and thus are more concerned with image generation. Interestingly, we clearly see that the register tokens encode a high degree of sequence-specific information, since we observe best performance for 2 of the 3 tasks when probing these tokens. This, together with the improved video prediction performance when including register tokens as reported in Table~\ref{tab:2}, highlights the value of including register tokens to the input.

Finally, we include probing tests on out-of-distribution parameter ranges not seen during training (details of these ranges can be found in Appendix \ref{App:1}). We see a small increase in MAE averaged over each dataset, from 0.17 (in-distribution) to 0.21 (out-of-distribution), suggesting strong generalisation of the learned PDE dynamics. 

\begin{table}[ht!]
    \centering
    \caption{\noindent Parameter estimation probing results for PSViT-small}
    \label{tab:4}
    \resizebox{0.48\textwidth}{!}{%
    \begin{tabular}{l l c c c c}
        \toprule
        && \multicolumn{4}{c}{Regression Error (MAE) $\downarrow$} \\
        \noalign{\smallskip}
        \cline{3-6}
        \noalign{\smallskip}
        Model Depth && \multicolumn{1}{c}{Roller (\textit{G})} & \multicolumn{1}{c}{Pendulum (\textit{G})}  & \multicolumn{1}{c}{Moon (\textit{M})} & \multicolumn{1}{c}{Average} \\
        \midrule
        Layer 1  && 0.47 & 0.52 & 0.63 & 0.54 \\
        Layer 2  && 0.21 & 0.34 & 0.45 & 0.33 \\
        Layer 3  && 0.20 & 0.31 & 0.42 & 0.31 \\
        Layer 4  && \textbf{0.16} & 0.27 & 0.32 & 0.25 \\
        Layer 5  && 0.17 & \textbf{0.24} & \textbf{0.29} & \textbf{0.23} \\
        Layer 6  && 0.21 & 0.30 & 0.33 & 0.28 \\
        Layer 7  && 0.35 & 0.41 & 0.47 & 0.41 \\
        Layer 8  && 0.62 & 0.78 & 0.80 & 0.73 \\
        \midrule
        Concatenation            && 0.15 & 0.25 & 0.32 & 0.24 \\
        Scalar mixing            && 0.15 & 0.28 & 0.30 & 0.24 \\
        Registers                && \textbf{0.14} & 0.20 & \textbf{0.19} & \textbf{0.18} \\
        \midrule
        In-distribution best     && 0.14 & 0.18 & 0.19 & 0.17 \\
        Out-of-distribution best && 0.16 & 0.22 & 0.24 & 0.21 \\
        \midrule
        Baseline                 && 0.83 & 0.90 & 0.91 & 0.88 \\
        \bottomrule
    \end{tabular}}
\end{table}

\subsection{Limitations}
For three of the simulation datasets increasing PSViT model size has minimal impact on performance over time. We would need to conduct further experiments to show the impact of larger scale models. For stochastic video-generation results on BAIR, our approach falls short compared to the latent space models, suggesting higher quality image synthesis with these approaches. This is backed up by our qualitative findings that object shapes, although positioned accurately, can distort over time, particularly when rotation is involved. On a similar vein, by performing end-to-end training, we don't benefit from the large-scale pretrained image encoders/decoders often employed by latent space models. It remains to be seen if this technique can benefit from scale in this manner.

\section{Conclusion and Future Work}

In conclusion, we presented a simple, interpretable pure transformer for end-to-end autoregressive video prediction. Our model utilises a novel U-Net style architecture and self-attention layout, removing the need for complex training strategies or latent feature-learning. It extends the horizon for physically accurate predictions by up to 50\% compared to existing approaches, while maintaining competitive performance on standard video quality metrics. Furthermore, we identified network regions correlated with specific spatiotemporal events and the encoding of sequence-specific PDE parameters. Through extensive experiments on physical simulation datasets using object tracking metrics, we demonstrated our model's proficiency in predicting PDE-driven video sequences. This model's interpretability, simplicity, and effectiveness highlight the benefits of a straightforward end-to-end approach.

Future work includes developing more sophisticated metrics for evaluating physical accuracy, moving beyond current object-based pixel distance methods. We also plan to train larger models on higher-resolution datasets featuring increased visual fidelity and physical complexity. This research paves the way for continued focus on simple yet effective spatiotemporal modelling and the development of accurate generative models for video content depicting physical systems.

\appendix
\label{App:1}

\section*{Dataset Details}
\begin{table}[ht!]
    \centering
    \caption{Dataset Specifications}
    \label{tab:a1}
    \resizebox{0.45\textwidth}{!}{%
    \begin{tabular}{l l c c c c c}
        \toprule
        Dataset && Samples & Seq. Length & Resolution & Num. Objects & PDE Variables \\
        \midrule
        Roller      && 5,000 & 100 & 64$\times$64   & 1     & 3 \\
        Pendulum    && 5,000 & 100 & 64$\times$64   & 1     & 5 \\
        Moon        && 5,000 & 100 & 64$\times$64   & 2     & 4 \\
        3D Balls    && 5,000 & 100 & 64$\times$64   & 1-3   & 1 \\
        CLEVRER*     && 5,000 & 140 & 128$\times$128 & 5     & - \\
        Fluid*       && 1,000 & 100 & 128$\times$128 & -     & - \\
        \bottomrule
    \end{tabular}}
\end{table}

Table~\ref{tab:a1} details the simulation datasets from the main work. Datasets used for parameter estimation probing include: the Roller dataset, where gravity is varied within an in-distribution range of 0-100 and an out-of-distribution range of 100-150; the Pendulum dataset, with gravity ranges of 0-6 (in-distribution) and 6-10 (out-of-distribution); and the Moon dataset, for which mass is varied between 72-200 (in-distribution) and 200-300 (out-of-distribution). These datasets are further described below.

The \textit{Moon} simulation consists of an \textit{orbiting moon}; treated as a rigid body with a random initial velocity, and a \textit{static celestial body}. The simulation models the gravitational interaction between the moon and the static celestial body as follows:

\begin{equation}
m_m \frac{d^2\vec{r}}{dt^2} = -\frac{G M_c m_m}{|\vec{r}|^3} \vec{r} ,
\label{eq13}
\end{equation}

\noindent where $\vec{r}$ is the position vector of the moon relative to the centre of the celestial body, $|\vec{r}|$ is the distance between the two bodies, $m_m$ is the mass of the moon, $M_c$ is the mass of the celestial body, $G$ is the gravitational constant, and $\frac{d^2\vec{r}}{dt^2}$ represents the acceleration of the moon due to gravity. For each sequence, we vary the initial tangential velocity of the moon, the radius of the moon, the radius of the celestial body, and the mass of the celestial body, simulating different orbital trajectories. $G$ and $m_m$ are kept constant for all samples.

The \textit{Pendulum} simulation consists of a single pendulum modelled as a point mass at the end of a massless rod. The pendulum swings about a fixed pivot point, which serves as the centre of rotation for all simulated sequences. The dynamics are governed by the following:
\begin{equation}
\theta''(t) = -\frac{G}{l} \sin(\theta(t)) ,
\label{eq14}
\end{equation}

\noindent where \(\theta(t)\) is the angle of the pendulum relative to the vertical, \(G\) represents the gravitational strength, and \(l\) is the length of the pendulum. For each sequence, we vary the following parameters: the \textit{initial angle} of the pendulum, \textit{gravitational strength}, \textit{pendulum length}, \textit{pendulum mass}, and the \textit{size of the pendulum}.

The \textit{Roller} simulation exhibits the motion of a ball of mass \( M \) rolling down a curved track under the influence of gravity. The force acting on the ball along the track is given by the equation:
\begin{equation}
F = M \cdot g \cdot \cos(\theta) ,
\label{eq15}
\end{equation}
where \( F \) is the force, \( M \) is the mass of the ball, \( G \) is the acceleration due to gravity, and \( \theta \) is the angle between the track and the horizontal plane. The ball transitions to free flight if the normal acceleration exceeds the limit set by the curve's radius of curvature at any point. This condition is mathematically represented as $a_n > \frac{v^2}{k}$, 
where \( a_n \) is the normal acceleration, \( v \) is the velocity of the ball, and \( k \) is the radius of curvature at the current point on the track. We vary the gravity strength \( g \) and the initial position.

\begin{figure}[b!]
    \centering 
    \includegraphics[width=0.49\textwidth]{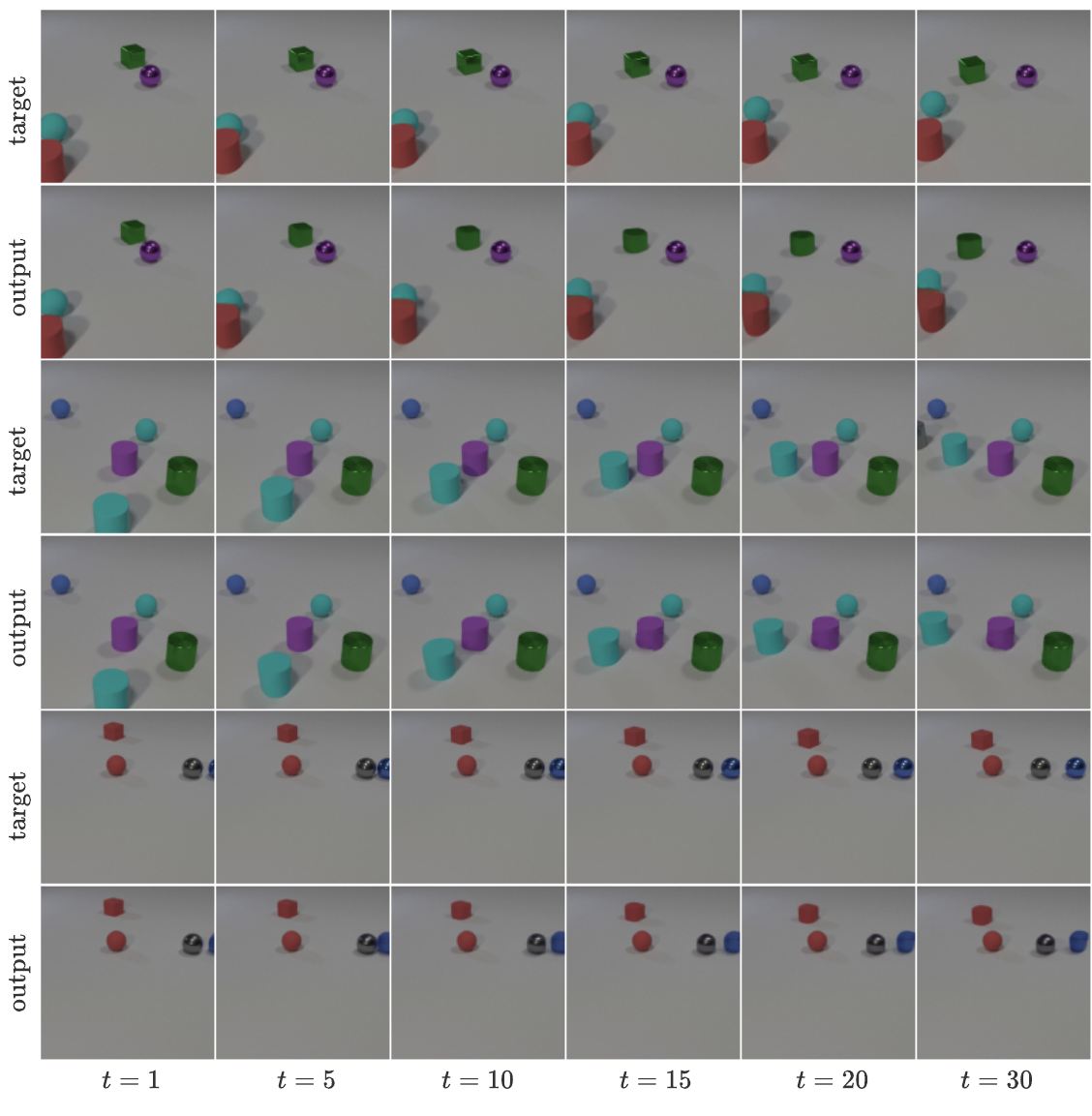}\\
    \caption{Example outputs from our PSViT model trained on the CLEVRER dataset, containing objects entering the scene partially or mostly obscured.}
    \label{figa2} 
\end{figure}
\begin{figure}[tb!]
    \centering 
    \includegraphics[width=0.49\textwidth]{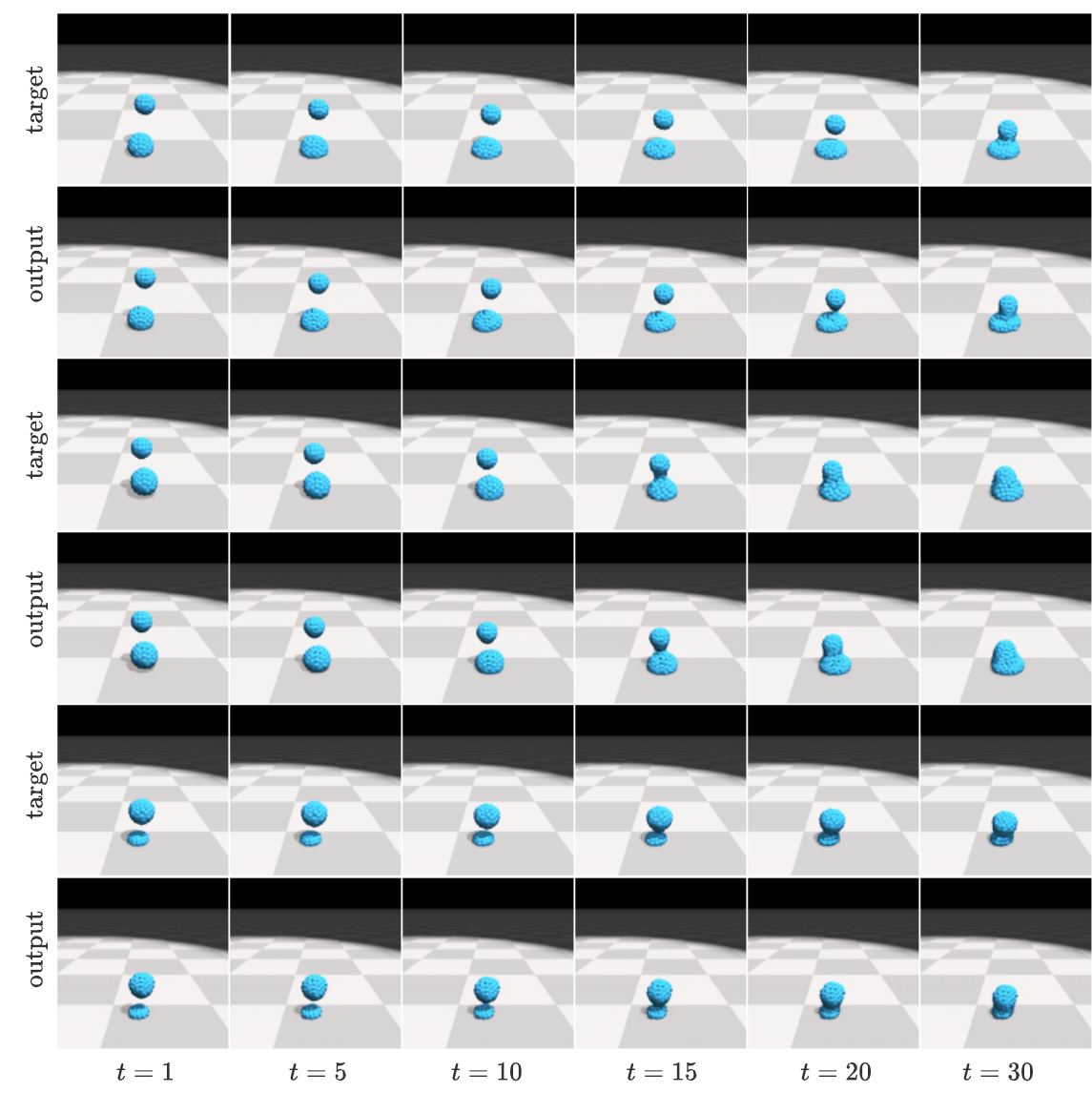}\\
    \caption{Example outputs from our PSViT model trained on the Fluid dataset. Examples shown are randomly sampled.}
    \label{figa3}
\end{figure}
\begin{figure}[t!]
    \centering 
    \includegraphics[width=0.49\textwidth]{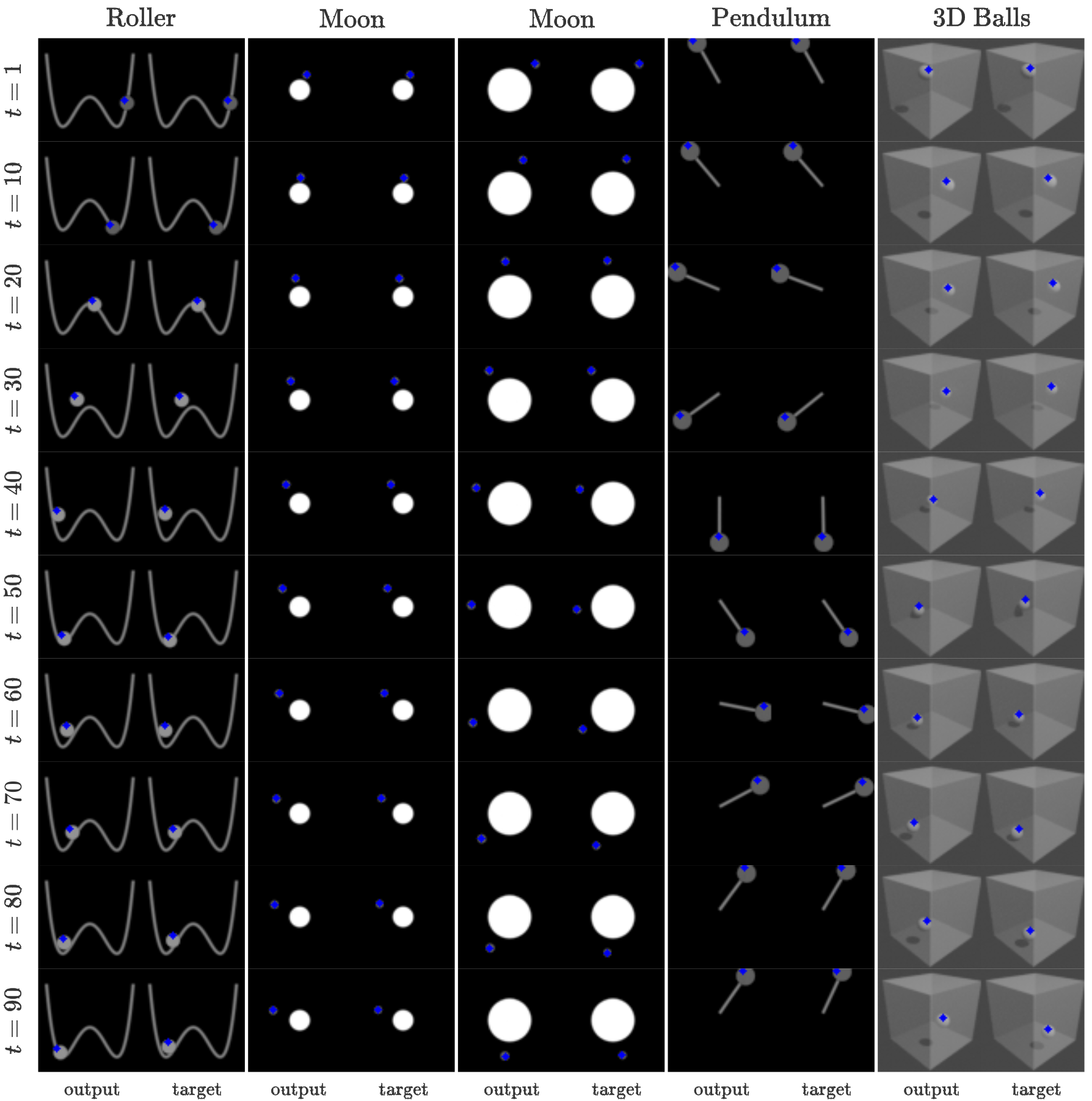}\\
    \caption{Example outputs from our PSViT model trained on the Roller, Moon, Pendulum, and 3D Balls datasets, annotated with object tracking positions. Examples shown are randomly sampled.}
    \label{figa1}
\end{figure}

\section*{Qualitative Outputs}
Fig.~\ref{figa2} shows CLEVRER outputs with objects appearing from out of view; while object positions may be accurate, rotation is often poorly modelled, especially post-collision. Fig.~\ref{figa3} contains Fluid dataset examples, and Fig.~\ref{figa1} shows Roller, Moon, Pendulum, and 3D Balls examples.

\section*{PSViT Efficiency}\label{app:efficiency}
\begin{table}[h!]
    \centering
    \caption{PSViT Efficiency Analysis}
    \label{tab:efficiency}
    \resizebox{0.99\linewidth}{!}{%
        \begin{tabular}{l c c c}
            \toprule
            Model Size & Resolution & \makecell[c]{Training Time (2000 epochs) \\ GPU Hours (Memory)} & \makecell[c]{Inference \\ FLOPs (Memory)}\\
            \midrule
            PSViT small  & $64 \times 64$ & $\approx52$ (33 GB)   & 14.1 G (509 MB)         \\ 
            PSViT small  & $128 \times 128$ & $\approx95$ (52 GB)  & $49.2$ G (1.5 GB)  \\ 
            \midrule
            PSViT medium & $64 \times 64$ & $\approx78$ (44 GB)   & $20.9$ G (731 MB)  \\ 
            PSViT medium & $128 \times 128$ & $\approx145$ (69 GB)  & $70.7$ G (1.99 GB)  \\
            \bottomrule
        \end{tabular}
    }
\end{table}
Table \ref{tab:efficiency} reports training and inference costs for each model size and two input resolutions. For inference, we report per-frame generation on a single 20 frame input sample.



\bibliographystyle{IEEEtranN}
\bibliography{finalbib}

\begin{thebibliography}{62}
\providecommand{\natexlab}[1]{#1}
\providecommand{\url}[1]{#1}
\csname url@samestyle\endcsname
\providecommand{\newblock}{\relax}
\providecommand{\bibinfo}[2]{#2}
\providecommand{\BIBentrySTDinterwordspacing}{\spaceskip=0pt\relax}
\providecommand{\BIBentryALTinterwordstretchfactor}{4}
\providecommand{\BIBentryALTinterwordspacing}{\spaceskip=\fontdimen2\font plus
\BIBentryALTinterwordstretchfactor\fontdimen3\font minus \fontdimen4\font\relax}
\providecommand{\BIBforeignlanguage}[2]{{%
\expandafter\ifx\csname l@#1\endcsname\relax
\typeout{** WARNING: IEEEtranN.bst: No hyphenation pattern has been}%
\typeout{** loaded for the language `#1'. Using the pattern for}%
\typeout{** the default language instead.}%
\else
\language=\csname l@#1\endcsname
\fi
#2}}
\providecommand{\BIBdecl}{\relax}
\BIBdecl

\bibitem[Vaswani et~al.(2017)Vaswani, Shazeer, Parmar, Uszkoreit, Jones, Gomez, Kaiser, and Polosukhin]{vaswani2017attention}
A.~Vaswani, N.~Shazeer, N.~Parmar, J.~Uszkoreit, L.~Jones, A.~N. Gomez, {\L}.~Kaiser, and I.~Polosukhin, ``Attention is all you need,'' \emph{Advances in neural information processing systems}, vol.~30, 2017.

\bibitem[Dosovitskiy et~al.(2021)Dosovitskiy, Beyer, Kolesnikov, Weissenborn, Zhai, Unterthiner, Dehghani, Minderer, Heigold, Gelly, Uszkoreit, and Houlsby]{dosovitskiy2021an}
A.~Dosovitskiy, L.~Beyer, A.~Kolesnikov, D.~Weissenborn, X.~Zhai, T.~Unterthiner, M.~Dehghani, M.~Minderer, G.~Heigold, S.~Gelly, J.~Uszkoreit, and N.~Houlsby, ``An image is worth 16x16 words: Transformers for image recognition at scale,'' in \emph{International Conference on Learning Representations}, 2021.

\bibitem[Yan et~al.(2021)Yan, Zhang, Abbeel, and Srinivas]{yan2021videogpt}
W.~Yan, Y.~Zhang, P.~Abbeel, and A.~Srinivas, ``Videogpt: Video generation using vq-vae and transformers,'' 2021.

\bibitem[Oprea et~al.(2020)Oprea, Martinez-Gonzalez, Garcia-Garcia, Castro-Vargas, Orts-Escolano, Garcia-Rodriguez, and Argyros]{oprea2020review}
S.~Oprea, P.~Martinez-Gonzalez, A.~Garcia-Garcia, J.~A. Castro-Vargas, S.~Orts-Escolano, J.~Garcia-Rodriguez, and A.~Argyros, ``A review on deep learning techniques for video prediction,'' \emph{IEEE Transactions on Pattern Analysis and Machine Intelligence}, vol.~44, no.~6, pp. 2806--2826, 2020.

\bibitem[Razavi et~al.(2019)Razavi, Van~den Oord, and Vinyals]{razavi2019generating}
A.~Razavi, A.~Van~den Oord, and O.~Vinyals, ``Generating diverse high-fidelity images with vq-vae-2,'' \emph{Advances in neural information processing systems}, vol.~32, 2019.

\bibitem[Zhang et~al.(2023)Zhang, Zhang, Zhang, and Kweon]{zhang2023text}
C.~Zhang, C.~Zhang, M.~Zhang, and I.~S. Kweon, ``Text-to-image diffusion models in generative ai: A survey,'' \emph{arXiv preprint arXiv:2303.07909}, 2023.

\bibitem[Castrejon et~al.(2019)Castrejon, Ballas, and Courville]{castrejon2019improved}
L.~Castrejon, N.~Ballas, and A.~Courville, ``Improved conditional vrnns for video prediction,'' in \emph{Proceedings of the IEEE/CVF international conference on computer vision}, 2019, pp. 7608--7617.

\bibitem[Zhou et~al.(2020)Zhou, Dong, and El~Saddik]{zhou2020deep}
Y.~Zhou, H.~Dong, and A.~El~Saddik, ``Deep learning in next-frame prediction: A benchmark review,'' \emph{IEEE Access}, vol.~8, pp. 69\,273--69\,283, 2020.

\bibitem[Wei et~al.(2022)Wei, Tay, Bommasani, Raffel, Zoph, Borgeaud, Yogatama, Bosma, Zhou, Metzler, Chi, Hashimoto, Vinyals, Liang, Dean, and Fedus]{wei2022emergent}
J.~Wei, Y.~Tay, R.~Bommasani, C.~Raffel, B.~Zoph, S.~Borgeaud, D.~Yogatama, M.~Bosma, D.~Zhou, D.~Metzler, E.~H. Chi, T.~Hashimoto, O.~Vinyals, P.~Liang, J.~Dean, and W.~Fedus, ``Emergent abilities of large language models,'' \emph{Transactions on Machine Learning Research}, 2022, survey Certification.

\bibitem[Brown et~al.(2020)Brown, Mann, Ryder, et~al.]{Brown2020LanguageMA}
T.~Brown, B.~Mann, N.~Ryder \emph{et~al.}, ``Language models are few-shot learners,'' in \emph{NeurIPS}, vol.~33, 2020, pp. 1877--1901.

\bibitem[Kohl et~al.(2023)Kohl, Chen, and Thuerey]{kohl2023_acdm}
G.~Kohl, L.~Chen, and N.~Thuerey, ``Benchmarking autoregressive conditional diffusion models for turbulent flow simulation,'' \emph{arXiv}, 2023.

\bibitem[Sønderby et~al.(2020)Sønderby, Espeholt, Heek, Dehghani, Oliver, Salimans, Agrawal, Hickey, and Kalchbrenner]{sønderby2020metnet}
C.~K. Sønderby, L.~Espeholt, J.~Heek, M.~Dehghani, A.~Oliver, T.~Salimans, S.~Agrawal, J.~Hickey, and N.~Kalchbrenner, ``Metnet: A neural weather model for precipitation forecasting,'' 2020.

\bibitem[Finn and Levine(2017)]{7989324}
C.~Finn and S.~Levine, ``Deep visual foresight for planning robot motion,'' in \emph{2017 IEEE International Conference on Robotics and Automation (ICRA)}, 2017, pp. 2786--2793.

\bibitem[Wen et~al.(2023)Wen, Zhao, Liu, Jia, Wang, Luo, Zhang, Wang, Sun, and Zhang]{wen2023panacea}
Y.~Wen, Y.~Zhao, Y.~Liu, F.~Jia, Y.~Wang, C.~Luo, C.~Zhang, T.~Wang, X.~Sun, and X.~Zhang, ``Panacea: Panoramic and controllable video generation for autonomous driving,'' 2023.

\bibitem[Hu et~al.(2023)Hu, Russell, Yeo, Murez, Fedoseev, Kendall, Shotton, and Corrado]{hu2023gaia1generativeworldmodel}
A.~Hu, L.~Russell, H.~Yeo, Z.~Murez, G.~Fedoseev, A.~Kendall, J.~Shotton, and G.~Corrado, ``Gaia-1: A generative world model for autonomous driving,'' 2023.

\bibitem[Gao et~al.(2022)Gao, Tan, Wu, and Li]{Gao_2022_CVPR}
Z.~Gao, C.~Tan, L.~Wu, and S.~Z. Li, ``Simvp: Simpler yet better video prediction,'' in \emph{Proceedings of the IEEE/CVF Conference on Computer Vision and Pattern Recognition (CVPR)}, June 2022, pp. 3170--3180.

\bibitem[Xing et~al.(2023)Xing, Feng, Chen, Dai, Hu, Xu, Wu, and Jiang]{xing2023survey}
Z.~Xing, Q.~Feng, H.~Chen, Q.~Dai, H.~Hu, H.~Xu, Z.~Wu, and Y.-G. Jiang, ``A survey on video diffusion models,'' 2023.

\bibitem[Croitoru et~al.(2023)Croitoru, Hondru, Ionescu, and Shah]{croitoru2023diffusion}
F.-A. Croitoru, V.~Hondru, R.~T. Ionescu, and M.~Shah, ``Diffusion models in vision: A survey,'' \emph{IEEE Transactions on Pattern Analysis and Machine Intelligence}, vol.~45, no.~9, pp. 10\,850--10\,869, 2023.

\bibitem[Yu et~al.(2023)Yu, Cheng, Sohn, Lezama, Zhang, Chang, Hauptmann, Yang, Hao, Essa, et~al.]{yu2023magvit}
L.~Yu, Y.~Cheng, K.~Sohn, J.~Lezama, H.~Zhang, H.~Chang, A.~G. Hauptmann, M.-H. Yang, Y.~Hao, I.~Essa \emph{et~al.}, ``Magvit: Masked generative video transformer,'' in \emph{Proceedings of the IEEE/CVF Conference on Computer Vision and Pattern Recognition}, 2023, pp. 10\,459--10\,469.

\bibitem[Ming et~al.(2024)Ming, Huang, Ju, Hu, Peng, and Zhou]{ming2024survey}
R.~Ming, Z.~Huang, Z.~Ju, J.~Hu, L.~Peng, and S.~Zhou, ``A survey on video prediction: From deterministic to generative approaches,'' \emph{arXiv preprint arXiv:2401.14718}, 2024.

\bibitem[van~den Oord et~al.(2018)van~den Oord, Li, and Vinyals]{Oord2018RepresentationLW}
A.~van~den Oord, Y.~Li, and O.~Vinyals, ``Representation learning with contrastive predictive coding,'' \emph{ArXiv}, vol. abs/1807.03748, 2018.

\bibitem[Donahue and Simonyan(2019)]{Donahue2019LargeSA}
J.~Donahue and K.~Simonyan, ``Large scale adversarial representation learning,'' in \emph{NeurIPS}, 2019.

\bibitem[Ranzato et~al.(2014)Ranzato, Szlam, Bruna, Mathieu, Collobert, and Chopra]{ranzato2014video}
M.~Ranzato, A.~Szlam, J.~Bruna, M.~Mathieu, R.~Collobert, and S.~Chopra, ``Video (language) modeling: a baseline for generative models of natural videos,'' \emph{arXiv preprint arXiv:1412.6604}, 2014.

\bibitem[Chen et~al.(2020)Chen, Radford, Wu, Jun, Dhariwal, Luan, and Sutskever]{Chen2020GenerativePF}
M.~Chen, A.~Radford, J.~Wu, H.~Jun, P.~Dhariwal, D.~Luan, and I.~Sutskever, ``Generative pretraining from pixels,'' in \emph{ICML}, 2020.

\bibitem[van~den Oord et~al.(2017)van~den Oord, Vinyals, and Kavukcuoglu]{Oord2017NeuralDR}
A.~van~den Oord, O.~Vinyals, and K.~Kavukcuoglu, ``Neural discrete representation learning,'' in \emph{NIPS}, 2017.

\bibitem[Wu et~al.(2022)Wu, Liang, Ji, Yang, Fang, Jiang, and Duan]{wu2022nuwa}
C.~Wu, J.~Liang, L.~Ji, F.~Yang, Y.~Fang, D.~Jiang, and N.~Duan, ``N{\"u}wa: Visual synthesis pre-training for neural visual world creation,'' in \emph{European conference on computer vision}.\hskip 1em plus 0.5em minus 0.4em\relax Springer, 2022, pp. 720--736.

\bibitem[Hong et~al.(2023)Hong, Ding, Zheng, Liu, and Tang]{hong2023cogvideo}
W.~Hong, M.~Ding, W.~Zheng, X.~Liu, and J.~Tang, ``Cogvideo: Large-scale pretraining for text-to-video generation via transformers,'' in \emph{The Eleventh International Conference on Learning Representations}, 2023.

\bibitem[Dosovitskiy and Brox(2016)]{Dosovitskiy2016GeneratingIW}
A.~Dosovitskiy and T.~Brox, ``Generating images with perceptual similarity metrics based on deep networks,'' in \emph{NIPS}, 2016.

\bibitem[van Amersfoort et~al.(2017)van Amersfoort, Kannan, Ranzato, Szlam, Tran, and Chintala]{Amersfoort2017TransformationBasedMO}
J.~R. van Amersfoort, A.~Kannan, M.~Ranzato, A.~Szlam, D.~Tran, and S.~Chintala, ``Transformation-based models of video sequences,'' \emph{ArXiv}, vol. abs/1701.08435, 2017.

\bibitem[Wang et~al.(2020)Wang, Wu, Long, and Tenenbaum]{wang2020probabilistic}
Y.~Wang, J.~Wu, M.~Long, and J.~B. Tenenbaum, ``Probabilistic video prediction from noisy data with a posterior confidence,'' in \emph{Proceedings of the IEEE/CVF Conference on Computer Vision and Pattern Recognition}, 2020, pp. 10\,830--10\,839.

\bibitem[Xie et~al.(2021)Xie, Wang, Yu, Anandkumar, Alvarez, and Luo]{xie2021segformer}
E.~Xie, W.~Wang, Z.~Yu, A.~Anandkumar, J.~M. Alvarez, and P.~Luo, ``Segformer: Simple and efficient design for semantic segmentation with transformers,'' in \emph{Neural Information Processing Systems (NeurIPS)}, 2021.

\bibitem[Bertasius et~al.(2021)Bertasius, Wang, and Torresani]{bertasius2021space}
G.~Bertasius, H.~Wang, and L.~Torresani, ``Is space-time attention all you need for video understanding?'' in \emph{ICML}, vol.~2, no.~3, 2021, p.~4.

\bibitem[Seo et~al.(2022)Seo, Lee, Liu, James, and Abbeel]{seo2022harp}
Y.~Seo, K.~Lee, F.~Liu, S.~James, and P.~Abbeel, ``Harp: Autoregressive latent video prediction with high-fidelity image generator,'' in \emph{2022 IEEE International Conference on Image Processing (ICIP)}.\hskip 1em plus 0.5em minus 0.4em\relax IEEE, 2022, pp. 3943--3947.

\bibitem[Weissenborn et~al.(2020)Weissenborn, Täckström, and Uszkoreit]{Weissenborn2020Scaling}
D.~Weissenborn, O.~Täckström, and J.~Uszkoreit, ``Scaling autoregressive video models,'' in \emph{International Conference on Learning Representations}, 2020.

\bibitem[Van Den~Oord et~al.(2016)Van Den~Oord, Kalchbrenner, and Kavukcuoglu]{10.5555/3045390.3045575}
A.~Van Den~Oord, N.~Kalchbrenner, and K.~Kavukcuoglu, ``Pixel recurrent neural networks,'' in \emph{Proceedings of the 33rd International Conference on International Conference on Machine Learning - Volume 48}, ser. ICML'16.\hskip 1em plus 0.5em minus 0.4em\relax JMLR.org, 2016, p. 1747–1756.

\bibitem[Rakhimov et~al.(2021)Rakhimov, Volkhonskiy, Artemov, Zorin, and Burnaev]{Rakhimov2021LatentVT}
R.~Rakhimov, D.~Volkhonskiy, A.~Artemov, D.~Zorin, and E.~Burnaev, ``Latent video transformer,'' in \emph{VISIGRAPP}, 2021.

\bibitem[Seo et~al.(2023)Seo, Lee, Kim, and Seo]{seo2023implicitstackedautoregressivemodel}
M.~Seo, H.~Lee, D.~Kim, and J.~Seo, ``Implicit stacked autoregressive model for video prediction,'' 2023.

\bibitem[Finn et~al.(2016)Finn, Goodfellow, and Levine]{10.5555/3157096.3157104}
C.~Finn, I.~Goodfellow, and S.~Levine, ``Unsupervised learning for physical interaction through video prediction,'' in \emph{Proceedings of the 30th International Conference on Neural Information Processing Systems}, ser. NIPS'16.\hskip 1em plus 0.5em minus 0.4em\relax Red Hook, NY, USA: Curran Associates Inc., 2016, p. 64–72.

\bibitem[Le~Guen and Thome(2020)]{9156800}
V.~Le~Guen and N.~Thome, ``Disentangling physical dynamics from unknown factors for unsupervised video prediction,'' in \emph{2020 IEEE/CVF Conference on Computer Vision and Pattern Recognition (CVPR)}, 2020, pp. 11\,471--11\,481.

\bibitem[Wu et~al.(2016)Wu, Lim, Zhang, Tenenbaum, and Freeman]{Wu2016Physics1L}
J.~Wu, J.~J. Lim, H.~Zhang, J.~B. Tenenbaum, and W.~T. Freeman, ``Physics 101: Learning physical object properties from unlabeled videos,'' in \emph{BMVC}, 2016.

\bibitem[Winterbottom et~al.(2024)Winterbottom, Hudson, Kluvanec, Slack, Sterling, Shentu, Xiao, Zhou, and Moubayed]{winterbottom2024powernextframepredictionlearning}
T.~Winterbottom, G.~T. Hudson, D.~Kluvanec, D.~Slack, J.~Sterling, J.~Shentu, C.~Xiao, Z.~Zhou, and N.~A. Moubayed, ``The power of next-frame prediction for learning physical laws,'' 2024.

\bibitem[Yin et~al.(2023)Yin, Kirchmeyer, Franceschi, Rakotomamonjy, and patrick gallinari]{yin2023continuous}
Y.~Yin, M.~Kirchmeyer, J.-Y. Franceschi, A.~Rakotomamonjy, and patrick gallinari, ``Continuous {PDE} dynamics forecasting with implicit neural representations,'' in \emph{The Eleventh International Conference on Learning Representations}, 2023.

\bibitem[Tompson et~al.(2017)Tompson, Schlachter, Sprechmann, and Perlin]{10.5555/3305890.3306035}
J.~Tompson, K.~Schlachter, P.~Sprechmann, and K.~Perlin, ``Accelerating eulerian fluid simulation with convolutional networks,'' in \emph{ICML}, 2017, p. 3424–3433.

\bibitem[Kolter and Manek(2019)]{kolter2019learning}
J.~Z. Kolter and G.~Manek, ``Learning stable deep dynamics models,'' \emph{Advances in neural information processing systems}, vol.~32, 2019.

\bibitem[de~Bezenac et~al.(2018)de~Bezenac, Pajot, and Gallinari]{de2018deep}
E.~de~Bezenac, A.~Pajot, and P.~Gallinari, ``Deep learning for physical processes: Incorporating prior scientific knowledge,'' in \emph{ICLR}, 2018.

\bibitem[Li et~al.(2021)Li, Kovachki, Azizzadenesheli, liu, Bhattacharya, Stuart, and Anandkumar]{li2021fourier}
Z.~Li, N.~B. Kovachki, K.~Azizzadenesheli, B.~liu, K.~Bhattacharya, A.~Stuart, and A.~Anandkumar, ``Fourier neural operator for parametric partial differential equations,'' in \emph{ICLR}, 2021.

\bibitem[Wang et~al.(2022)Wang, Walters, and Yu]{wang2022meta}
R.~Wang, R.~Walters, and R.~Yu, ``Meta-learning dynamics forecasting using task inference,'' \emph{Advances in Neural Information Processing Systems}, vol.~35, pp. 21\,640--21\,653, 2022.

\bibitem[Hendrycks and Gimpel(2016)]{hendrycks2016gaussian}
D.~Hendrycks and K.~Gimpel, ``Gaussian error linear units (gelus),'' \emph{arXiv preprint arXiv:1606.08415}, 2016.

\bibitem[Darcet et~al.(2024)Darcet, Oquab, Mairal, and Bojanowski]{darcet2024vision}
T.~Darcet, M.~Oquab, J.~Mairal, and P.~Bojanowski, ``Vision transformers need registers,'' in \emph{The Twelfth International Conference on Learning Representations}, 2024.

\bibitem[Butcher(1996)]{butcher1996history}
J.~C. Butcher, ``A history of runge-kutta methods,'' \emph{Applied numerical mathematics}, vol.~20, no.~3, pp. 247--260, 1996.

\bibitem[Yang et~al.(2019)Yang, Yang, Liu, Xiao, Davis, and Kautz]{Yang_2019_CVPR}
X.~Yang, X.~Yang, M.-Y. Liu, F.~Xiao, L.~S. Davis, and J.~Kautz, ``Step: Spatio-temporal progressive learning for video action detection,'' in \emph{CVPR}, 2019.

\bibitem[Li et~al.(2019)Li, Wu, Tedrake, Tenenbaum, and Torralba]{li2019learning}
Y.~Li, J.~Wu, R.~Tedrake, J.~B. Tenenbaum, and A.~Torralba, ``Learning particle dynamics for manipulating rigid bodies, deformable objects, and fluids,'' in \emph{ICLR}, 2019.

\bibitem[Srivastava et~al.(2015)Srivastava, Mansimov, and Salakhutdinov]{10.5555/3045118.3045209}
N.~Srivastava, E.~Mansimov, and R.~Salakhutdinov, ``Unsupervised learning of video representations using lstms,'' in \emph{Proceedings of the 32nd International Conference on International Conference on Machine Learning - Volume 37}, ser. ICML'15.\hskip 1em plus 0.5em minus 0.4em\relax JMLR.org, 2015, p. 843–852.

\bibitem[Ebert et~al.(2017)Ebert, Finn, Lee, and Levine]{ebert2017self}
F.~Ebert, C.~Finn, A.~X. Lee, and S.~Levine, ``Self-supervised visual planning with temporal skip connections.'' \emph{CoRL}, vol.~12, no.~16, p.~23, 2017.

\bibitem[Kingma and Ba(2015)]{KingBa15}
D.~Kingma and J.~Ba, ``Adam: A method for stochastic optimization,'' in \emph{International Conference on Learning Representations (ICLR)}, San Diega, CA, USA, 2015.

\bibitem[Wang et~al.(2004)Wang, Bovik, Sheikh, and Simoncelli]{1284395}
Z.~Wang, A.~Bovik, H.~Sheikh, and E.~Simoncelli, ``Image quality assessment: from error visibility to structural similarity,'' \emph{IEEE Transactions on Image Processing}, vol.~13, no.~4, pp. 600--612, 2004.

\bibitem[Zheng et~al.(2024)Zheng, Peng, Yang, Shen, Li, Liu, Zhou, Li, and You]{zheng2024open}
Z.~Zheng, X.~Peng, T.~Yang, C.~Shen, S.~Li, H.~Liu, Y.~Zhou, T.~Li, and Y.~You, ``Open-sora: Democratizing efficient video production for all,'' \emph{arXiv preprint arXiv:2412.20404}, 2024.

\bibitem[Zhao et~al.(2024)Zhao, Zhang, Cun, Yang, Niu, Li, Hu, and Shan]{zhao2024cvvae}
S.~Zhao, Y.~Zhang, X.~Cun, S.~Yang, M.~Niu, X.~Li, W.~Hu, and Y.~Shan, ``{CV}-{VAE}: A compatible video {VAE} for latent generative video models,'' in \emph{The Thirty-eighth Annual Conference on Neural Information Processing Systems}, 2024.

\bibitem[Zhang et~al.(2022)Zhang, Roller, Goyal, Artetxe, Chen, Chen, Dewan, Diab, Li, Lin, et~al.]{zhang2022opt}
S.~Zhang, S.~Roller, N.~Goyal, M.~Artetxe, M.~Chen, S.~Chen, C.~Dewan, M.~Diab, X.~Li, X.~V. Lin \emph{et~al.}, ``Opt: Open pre-trained transformer language models,'' \emph{arXiv preprint arXiv:2205.01068}, 2022.

\bibitem[Raffel et~al.(2020)Raffel, Shazeer, Roberts, Lee, Narang, Matena, Zhou, Li, and Liu]{raffel2020exploring}
C.~Raffel, N.~Shazeer, A.~Roberts, K.~Lee, S.~Narang, M.~Matena, Y.~Zhou, W.~Li, and P.~J. Liu, ``Exploring the limits of transfer learning with a unified text-to-text transformer,'' \emph{Journal of machine learning research}, vol.~21, no. 140, pp. 1--67, 2020.

\bibitem[Chowdhery et~al.(2023)Chowdhery, Narang, Devlin, Bosma, Mishra, Roberts, Barham, Chung, Sutton, Gehrmann, et~al.]{chowdhery2023palm}
A.~Chowdhery, S.~Narang, J.~Devlin, M.~Bosma, G.~Mishra, A.~Roberts, P.~Barham, H.~W. Chung, C.~Sutton, S.~Gehrmann \emph{et~al.}, ``Palm: Scaling language modeling with pathways,'' \emph{Journal of Machine Learning Research}, vol.~24, no. 240, pp. 1--113, 2023.

\bibitem[Unterthiner et~al.(2019)Unterthiner, van Steenkiste, Kurach, Marinier, Michalski, and Gelly]{Unterthiner2019FVDAN}
T.~Unterthiner, S.~van Steenkiste, K.~Kurach, R.~Marinier, M.~Michalski, and S.~Gelly, ``Fvd: A new metric for video generation,'' in \emph{DGS@ICLR}, 2019.

\end{thebibliography}

\end{document}